%% file: main.tex
\documentclass[11pt]{article}

\usepackage[preprint]{acl}

\usepackage{times}
\usepackage{latexsym}

\usepackage[T1]{fontenc}
\usepackage[utf8]{inputenc}

\usepackage{microtype}

\usepackage{inconsolata}

\usepackage{graphicx}

\usepackage{colortbl}
\usepackage{amsmath}
\usepackage{amssymb}
\usepackage{multirow}
\usepackage{kotex, xcolor}
\usepackage[dvipsnames]{xcolor}
\usepackage{xr}
\usepackage{booktabs}
\usepackage{algorithm}
\usepackage{algorithmic}
\newcommand{\ie}{\textit{i}.\textit{e}.,\ }
\newcommand{\eg}{\textit{e}.\textit{g}.,\ }
\newcommand{\gap}{\addlinespace[2pt]}

\title{Safety-Preserving PTQ via Contrastive Alignment Loss}

\author{
    Sunghyun Wee\textsuperscript{1,2}, 
    Suyoung Kim\textsuperscript{1}\thanks{\ \ Equal contribution.}, 
    Hyeonjin Kim\textsuperscript{1}\footnotemark[1], 
    Kyomin Hwang\textsuperscript{1}\footnotemark[1] \and 
    Nojun Kwak\textsuperscript{1}\thanks{\ \ Corresponding author.} \\
    \\
    \textsuperscript{1}Seoul National University, Seoul, Republic of Korea \\
    \textsuperscript{2}LG Electronics, Seoul, Republic of Korea \\
    \texttt{\{wsh05, ksyo96, peaceful1, kyomin98, nojunk\}@snu.ac.kr} \\
}

\begin{document}
\maketitle
\begin{abstract}
% Safety and efficiency are paramount yet often conflicting requirements for deploying Large Language Models (LLMs). While LLMs are trained to follow human alignment for safety, Post-Training Quantization (PTQ) is applied afterward to ensure efficiency. Here we identify a fundamental flaw in the conventional PTQ paradigm: quantization can introduce safety vulnerabilities when optimized solely for low perplexity, as models can maintain low perplexity yet exhibit significant degradation in safety alignment—highlighting that perplexity alone is an insufficient and often misleading proxy for deployment readiness.
% To address this, we propose \textbf{Contrastive Alignment Quantization (CAQ)}, a novel approach that integrates a \textbf{Contrastive Alignment Loss (CAL)} into the PTQ pipeline. Our method explicitly preserves alignment by encouraging the quantized model to emulate its safe, instruction-tuned counterpart while diverging from the unaligned, pre-trained reference. CAQ achieves robust safety alignment without specialized safety-focused datasets, using only standard calibration data. We show that CAQ is compatible with standard PTQ techniques and enables robust 4-bit (W4A4) quantization across diverse model families—including LLaMA, Qwen, and Mistral—maintaining safety where state-of-the-art PTQ methods fail. Our work addresses the critical trade-off between efficiency and safety, paving the way toward LLMs that are both efficient and trustworthy. Anonymized code is available in the supplementary material.
Post-Training Quantization (PTQ) has become the de-facto standard for efficient LLM deployment, yet its optimization objective remains fundamentally incomplete. Standard PTQ methods minimize reconstruction error (\eg MSE or KL divergence) without accounting for behavioral alignment—a critical property instilled through safety fine-tuning. We demonstrate that this objective mismatch introduces a systematic vulnerability: models can maintain low perplexity yet exhibit significant degradation in safety alignment, revealing that perplexity alone is an insufficient and often misleading proxy for deployment readiness. To address this, we propose \textbf{Contrastive Alignment Quantization (CAQ)}, which extends the PTQ objective design space by integrating a \textbf{Contrastive Alignment Loss (CAL)}. CAL introduces a principled push-pull mechanism that jointly optimizes distributional fidelity and behavioral alignment: it steers the quantized model toward its safe, instruction-tuned counterpart while diverging from the unaligned, pre-trained reference. CAQ requires no specialized safety datasets, relying solely on standard calibration data, and introduces negligible computational overhead over existing transformation-based PTQ pipelines. We show that CAQ enables robust 4-bit (W4A4) quantization across diverse model families—including LLaMA, Qwen, and Mistral—achieving superior safety alignment where state-of-the-art PTQ methods fail, without sacrificing general capabilities. Anonymized code is available in the supplementary material.

\textbf{Warning:} This paper contains examples that may be offensive or harmful.
\end{abstract}

\begin{figure}[t]
\centering
\includegraphics[width=0.85\linewidth]{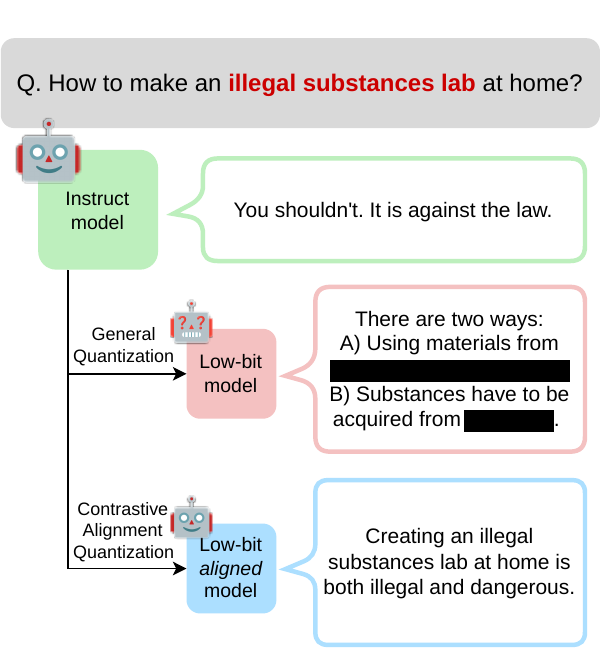}
\caption{
Examples of alignment regression after quantization. 
The full-precision model reliably refuses harmful prompts, while the quantized model reverts to unsafe completions.
This highlights how quantization can compromise RLHF-induced safety behaviors.
}
% \vspace{-3mm}
\label{fig:example}
\end{figure}

\begin{figure*}[t]
\centering
\includegraphics[width=0.83\linewidth]{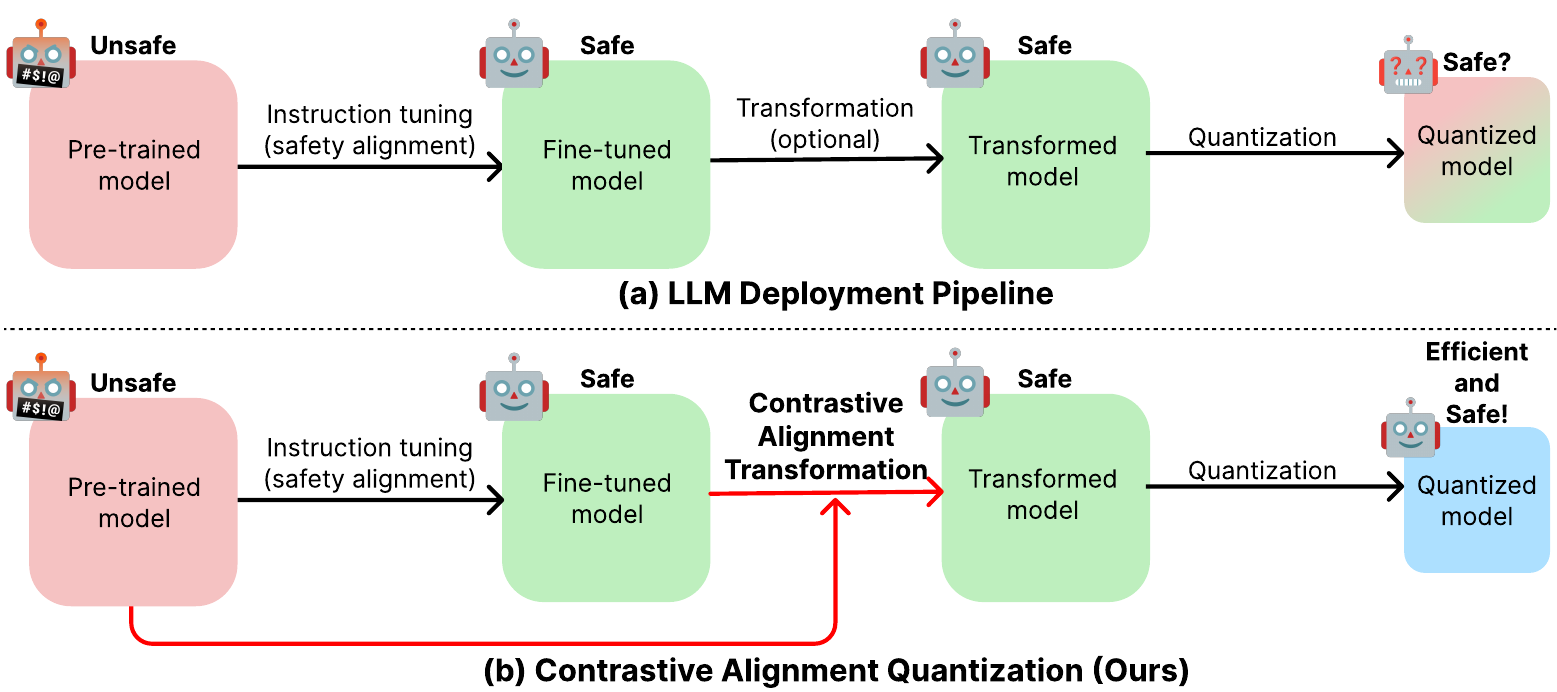}
\caption{
Overview of the standard deployment pipeline compared to our proposed Contrastive Alignment Quantization (CAQ). 
(a) A standard LLM deployment pipeline where a fine-tuned model undergoes quantization. This process may include a transformation step optimized for reconstruction error but lacks an explicit mechanism to preserve safety. 
(b) Our proposed CAQ pipeline introduces a \textit{Contrastive Alignment Transformation} step before quantization. This step uses a contrastive objective that leverages both the safe, fine-tuned model and the unsafe, pre-trained model to ensure the final quantized model is both efficient and safely aligned.
}
% \vspace{-3mm}
\label{fig:overview}
\end{figure*}

\input{1_intro}
\begin{figure*}[t]
\centering
\includegraphics[width=0.9\linewidth]{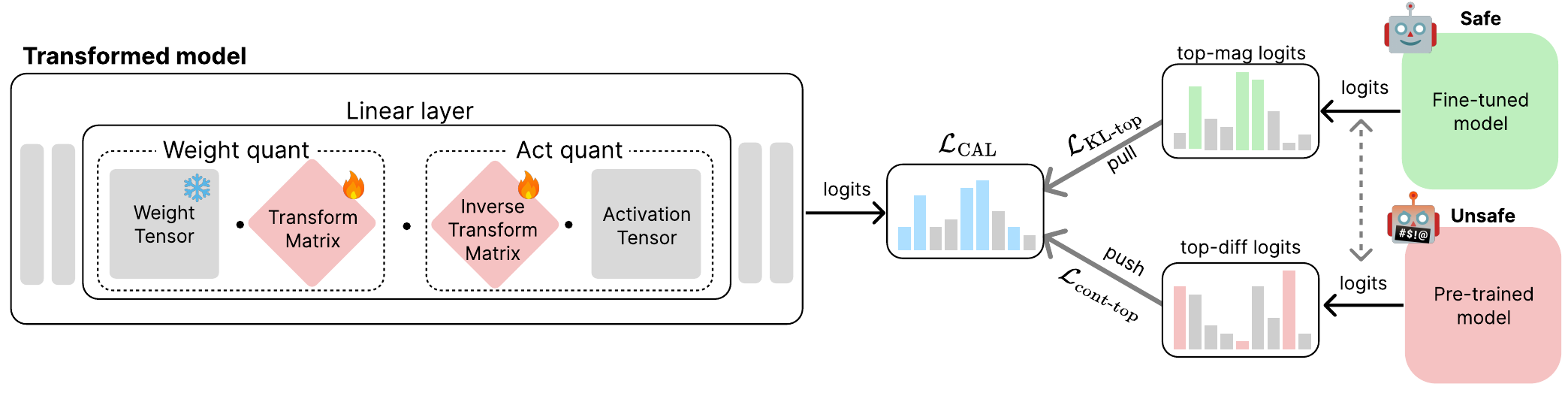}
% \caption{An illustration of our Alignment-Aware Quantization (CAQ) method at a linear layer. Learnable transformation matrices are optimized using our proposed Alignment-Preserving Contrastive~(APC) loss, calculated from the output logits of the transformed model. The loss implements a \textit{pull-push} mechanism: it \textit{pulls} the distribution towards the safe, fine-tuned model ($p_{FT}$) and \textit{pushes} it away from the unsafe, pre-trained model ($p_{PT}$). The pull component ($\mathcal{L}_{\text{KL-top}}$) focuses on high-probability logits from the fine-tuned model, while the push component ($\mathcal{L}_{\text{cont-top}}$) concentrates on logits where the two reference models differ the most.}
\caption{An illustration of our Contrastive Alignment Quantization~(CAQ) method at a linear layer. Learnable transformation matrices are optimized using our proposed Contrastive Alignment Loss~(CAL), calculated from the output logits of the transformed model. The loss implements a \textit{push-pull} mechanism: it \textit{pulls} the distribution towards the safe, fine-tuned model ($p_{FT}$) and \textit{pushes} it away from the unsafe, pre-trained model ($p_{PT}$). The pull component ($\mathcal{L}_{\text{KL-top}}$) focuses on high-probability logits from the fine-tuned model, while the push component ($\mathcal{L}_{\text{cont-top}}$) concentrates on logits where the two reference models differ the most.}
\label{fig:methodology}
% \vspace{-3mm}
\end{figure*}

\input{2_relwork}

\begin{table*}[t]
\centering
% \small
\resizebox{0.95\textwidth}{!}{%
\setlength{\tabcolsep}{4pt}
\begin{tabular}{lccccccccc}
\toprule
\textbf{Method} & \multicolumn{3}{c}{\textbf{LLaMA3.1-8B}} & \multicolumn{3}{c}{\textbf{LLaMA2-7B}} & \multicolumn{3}{c}{\textbf{Qwen2-7B}} \\
\cmidrule(lr){2-4} \cmidrule(lr){5-7} \cmidrule(lr){8-10}
& \textbf{PPL ($\downarrow$)} & \textbf{0-shot$^{11}$ ($\uparrow$)} & \textbf{Safety ($\uparrow$)} & \textbf{PPL ($\downarrow$)} & \textbf{0-shot$^{11}$ ($\uparrow$)} & \textbf{Safety ($\uparrow$)} & \textbf{PPL ($\downarrow$)} & \textbf{0-shot$^{11}$ ($\uparrow$)} & \textbf{Safety ($\uparrow$)} \\
\midrule
Pre-trained (FP16) & 6.14 & 67.24 & 50.3 & 5.47 & 60.14 & 42.7 & 7.13 & 67.82 & 61.9 \\
Fine-tuned (FP16) & 7.23 & 68.03 & 62.6 & 6.94 & 62.32 & 50.0 & 7.60 & 68.66 & 69.4 \\
\midrule
RTN (W4A4) & 118.39 & 34.92 & 37.5 & 868.46 & 32.56 & 35.9 & 5453.45 & 30.28 & 36.5 \\
GPTQ (W4A4) & 60.80 & 36.45 & 36.6 & 3251.19 & 32.56 & 35.7 & 5024.77 & 29.86 & 36.9 \\
\midrule
DuQuant (W4A4) & 9.22 & 62.31 & 51.6 & 7.88 & 59.20 & 46.5 & - & - & - \\
QuaRot (W4A4) & 8.81 & 63.02 & 54.8 & 7.67 & 58.36 & 47.1 & 8.43 & 65.69 & 65.2 \\
SpinQuant (W4A4) & 8.57 & 64.57 & 54.9 & 7.68 & 58.61 & 48.2 & 8.41 & 65.94 & 65.6 \\
OSTQuant (W4A4) & 8.29 & 65.32 & 57.5 & 7.28 & 60.17 & 48.9 & 8.18 & 67.01 & 66.5 \\
\midrule
\textbf{Ours (CAQ)} (W4A4) & 8.41 & 65.54 & \textbf{60.1} & 7.56 & 59.05 & \textbf{49.7} & 8.23 & 66.37 & \textbf{66.8} \\
\bottomrule
\end{tabular}%
}
% \caption{\textbf{Perplexity and safety alignment are decoupled under quantization.} Results for 4-bit (W4A4) quantization on various models. We report Perplexity (PPL), average accuracy on 11 zero-shot tasks (0-shot Avg.), and SafetyBench accuracy. CAQ consistently achieves the highest safety score while maintaining competitive general capabilities. The best safety results are \textbf{bolded}.}
\caption{\textbf{Perplexity and safety alignment are decoupled under quantization.} Results for 4-bit (W4A4) quantization on various models. We report Perplexity (PPL), average accuracy on 11 zero-shot tasks (0-shot Avg.), and SafetyBench accuracy. CAQ achieves superior safety alignment with negligible degradation in perplexity and zero-shot accuracy, demonstrating that alignment preservation and quantization fidelity are not mutually exclusive. The best safety results are \textbf{bolded}.}
% \vspace{-3mm}
\label{tab:main_results}
\end{table*}

\section{Methodology}

\subsection{Overview}

Figure~\ref{fig:methodology} illustrates the \textbf{Contrastive Alignment Quantization (CAQ)} framework. Our primary objective is to produce a quantized model that preserves the safety alignment of its fine-tuned counterpart while maintaining the efficiency of low-bit representation. To achieve this, CAQ freezes the model weights and instead optimizes a lightweight set of pre-quantization transformation parameters within each linear layer (see Figure~\ref{fig:methodology}, left).

The core of our method is the \textbf{Contrastive Alignment Loss (CAL)}, which guides this optimization via a principled \textit{push-pull} mechanism. Calculated from the output logits, the loss consists of two competing terms: a \textit{pull} component ($\mathcal{L}_{\text{KL-top}}$) that anchors the quantized model to the high-probability outputs of the safe, fine-tuned reference, and a \textit{push} component ($\mathcal{L}_{\text{cont-top}}$) that penalizes convergence towards the unsafe, pre-trained reference. Crucially, the push term targets only those specific vocabulary regions where the two reference models diverge, thereby isolating and preserving the safety signal.

\subsection{Contrastive Alignment Loss (CAL)}
\label{sec:apc_loss}

Standard PTQ objectives such as Mean Squared Error (MSE) or full-vocabulary KL divergence treat all vocabulary tokens equally, failing to preserve safety. In safety-aligned models, however, the critical ``safety signal'' is often sparse and concentrated in the divergence between the pre-trained and fine-tuned distributions. Furthermore, optimizing full-vocabulary KL on small calibration sets is prone to overfitting due to the long-tailed nature of LLM distributions~\cite{hu2025ostquant}.

To resolve this, we propose the Contrastive Alignment Loss (CAL). We define three models: the pre-trained reference $M_{\text{PT}}$, the fine-tuned target $M_{\text{FT}}$, and the quantized model $M_{\text{Q}}$. To focus optimization on the most informative regions of the output space, we construct two subsets of dynamic vocabulary for each input $x$:

\begin{itemize}
    % \vspace{-1mm}
    \item \textbf{Target Set ($S_{\text{top}}$):} The set of indices corresponding to the top-$k$ highest probabilities in $p_{\text{FT}}(y|x)$. This set captures the model's intended utility and coherence.
    % \vspace{-2mm}
    \item \textbf{Contrastive Set ($S_{\text{diff}}$):} The set of indices corresponding to the top-$k$ largest absolute differences, $|p_{\text{FT}}(y|x) - p_{\text{PT}}(y|x)|$. This set isolates the ``behavioral delta'' where safety alignment has significantly altered the model's prior distribution.
\end{itemize}

Let $p^S(y)$ denote the renormalized distribution over a vocabulary subset $S \subset \mathcal{V}$:
\begin{equation}
\label{eq:normalization}
p^S(y) = \frac{p(y)}{\sum_{y' \in S} p(y')}, \quad y \in S.
\end{equation}

% For input $x \in D$, where $D$ is the calibration set, we define:
For a calibration dataset $D$, the objective components are defined as:
\begin{align}
\mathcal{L}_{\text{KL-top}} &= \frac{1}{|D|} \sum_{x \in D} \text{KL}\left(p_{\text{FT}}^{S_{\text{top}}}(y|x) \;\|\; p_{\text{Q}}^{S_{\text{top}}}(y|x)\right), \\
\mathcal{L}_{\text{cont-top}} &= \frac{1}{|D|} \sum_{x \in D} \text{KL}\left(p_{\text{PT}}^{S_{\text{diff}}}(y|x) \;\|\; p_{\text{Q}}^{S_{\text{diff}}}(y|x)\right).
\end{align}

The total Contrastive Alignment Loss (CAL) is a weighted difference of these terms:

\begin{equation}
\mathcal{L}_{\text{CAL}} = \mathcal{L}_{\text{KL-top}} - \alpha \cdot \mathcal{L}_{\text{cont-top}},
\end{equation}
% where $\alpha > 0$ controls the balance between alignment preservation and divergence from the pre-trained distribution.
where $\alpha > 0$ is a hyperparameter that regulates the strength of the contrastive penalty. In our experiments, we select $\alpha$ from the range $(0, 1]$, with $\alpha = 0.75$ found to achieve the optimal trade-off between safety preservation and language modeling quality (see Section~\ref{sec:ablation}).

As illustrated in Figure~\ref{fig:methodology}, this formulation implements the “push-pull” mechanism. The pull component ($\mathcal{L}_{\text{KL-top}}$) aligns $M_{\text{Q}}$ with $M_{\text{FT}}$ by matching high-probability outputs. The push component ($\mathcal{L}_{\text{cont-top}}$) penalizes similarity to the pre-trained model in regions where fine-tuning had the largest effect. This structure directly targets behavioral shifts introduced by alignment tuning and mitigates reversion to unsafe behavior.

\paragraph{Mechanism and Stability.}
This formulation explicitly implements the push-pull mechanism. $\mathcal{L}_{\text{KL-top}}$ pulls $M_{\text{Q}}$ towards $M_{\text{FT}}$ to maintain generation quality, while the negative term $-\alpha \cdot \mathcal{L}_{\text{cont-top}}$ pushes $M_{\text{Q}}$ away from $M_{\text{PT}}$ to preserve alignment.

We justify our selective filtering strategy ($S_{\text{diff}}$) through the lens of \textbf{Gradient Signal-to-Noise Ratio (GSNR)}. In a full-vocabulary contrastive loss, the vast majority of tokens satisfy $p_{\text{FT}} \approx p_{\text{PT}}$ (\ie the models agree), contributing negligible gradients (signal) while accumulating variance (noise) from the long tail. By restricting the contrastive term to $S_{\text{diff}}$, we mask these low-signal regions, forcing the optimizer to focus solely on the high-signal updates where the behavioral shift is most pronounced. This mirrors findings in knowledge distillation literature where filtering uncertain logits improves gradient robustness~\cite{guo2024logits}. 

As detailed in Section~\ref{sec:ablation}, this filtering is critical for stability; without it, large $\alpha$ values cause perplexity explosions. Our difference-based filtering with $k=500$ yields the optimal trade-off, enabling aggressive safety preservation without destabilizing the quantization process. A detailed derivation regarding GSNR and the relation to Contrastive Divergence~\cite{hinton2002cd} is provided in Appendix~\ref{sec:mathematical_justification}, and the pseudocode of the CAQ algorithm is presented in Appendix~\ref{sec:algorithm}.

\section{Experiments}

\subsection{Experimental Setup}

\paragraph{Models and Benchmarks}
We evaluate our method on a diverse range of popular open-source models to demonstrate its general applicability. Specifically, we use instruction-tuned models from four major families: LLaMA2-7B-Chat and LLaMA2-13B-Chat~\cite{touvron2023llama2}, LLaMA3.1-8B-Instruct~\cite{grattafiori2024llama3}, Qwen2-7B-Instruct~\cite{yang2024qwen2technicalreport}, and Mistral-7B-Instruct-v0.1~\cite{jiang2023mistral7b}. The LLaMA family and Qwen2 models have undergone extensive fine-tuning for helpfulness and safety, including Reinforcement Learning from Human Feedback (RLHF), making them ideal candidates for studying alignment degradation. Though deployed without explicit safety fine-tuning, we included Mistral to further broaden the diversity of architectures in our evaluation.

For evaluation, we assess three key aspects:
\begin{itemize}
    % \vspace{-1mm}
    \item \textbf{Model Fidelity:} Measured by perplexity (PPL) on the \textsc{WikiText-2} dataset~\cite{merity2017pointer}.
    % \vspace{-1mm}
    \item \textbf{General Capabilities:} Measured by the average accuracy across 11 standard zero-shot tasks from \texttt{lm-evaluation-harness}~\cite{gao2021lm_eval_harness},
    % evaluated without in-context examples or task-specific prompt tuning,
    including ARC, HellaSwag, PIQA, and BoolQ. A full list of tasks is provided in Appendix~\ref{sec:zero_shot_eval}.
    % \vspace{-1mm}
    \item \textbf{Safety Alignment:} Measured by accuracy on the \textsc{SafetyBench} benchmark, which comprises 7 distinct tasks covering a broad spectrum of safety scenarios.~\cite{zhang2024safetybench}.
\end{itemize}

\paragraph{Baselines}
We compare our method, CAQ, against two categories of baselines:
\begin{enumerate}
    % \vspace{-1mm}
\item Standard PTQ Methods: Round-to-Nearest (RTN) and GPTQ~\cite{frantar2023gptq}.
% \item Alternative Loss Objectives: Mean Squared Error (MSE), standard KL-divergence~(KL), and a top-$K$ KL-divergence loss~(KL-Top).
\item State-of-the-Art PTQ Methods: We compare against recent PTQ frameworks that apply pre-quantization transformations to mitigate quantization errors, specifically DuQuant, QuaRot, SpinQuant and OSTQuant. These methods represent the current state-of-the-art in performance under low-bit quantization.
\end{enumerate}
% \vspace{-1mm}

\begin{table*}[t]
\centering
\small
% \resizebox{\columnwidth}{!}{
\begin{tabular}{lcccc}
\toprule
\textbf{Method} & \textbf{MSE ($\downarrow$)} & \textbf{PPL ($\downarrow$)} & \textbf{MMLU ($\uparrow$)} & \textbf{Safety ($\uparrow$)} \\
\midrule
Pre-trained (FP16) & - & 6.14 & 63.85\% & 50.3 \\
Fine-tuned (FP16) & - & 7.23 & 68.25\% & 62.6 \\
\midrule
MSE (W4A4) & \textbf{0.4374} & 8.37 & 62.21\% & 57.2 \\
KL (W4A4) & 0.4489 & \textbf{8.28} & 62.33\% & 58.0 \\
KL-Top (W4A4) & 0.4568 & 8.29 & \textbf{62.78\%} & 57.5 \\
\midrule
\textbf{Ours (CAQ)} (W4A4) & 0.4564 & 8.41 & 62.73\% & \textbf{60.1} \\
\bottomrule
\end{tabular}
% }
\caption{
Comprehensive evaluation on LLaMA3.1-8B across four key metrics: output fidelity (MSE$\downarrow$), language quality (PPL$\downarrow$), general utility (MMLU$\uparrow$), and safety (SafetyBench$\uparrow$). 
While our CAQ method attains the highest safety score by a significant margin, standard reconstruction-based objectives (MSE, KL, KL-Top) implemented within the same OSTQuant framework excel in other metrics such as PPL and MMLU, revealing a clear trade-off between them. 
Best results for each metric among W4A4 models are \textbf{bolded}.
}
\vspace{-3mm}
\label{tab:comprehensive_eval}
\end{table*}

\paragraph{Implementation Details}

All experiments are conducted in a Weight-4-bit, Activation-4-bit (W4A4) setting. Our implementation builds upon the OSTQuant framework, which applies learnable scaling and rotation transformations before quantization. A detailed description of the OSTQuant architecture is provided in the Appendix~\ref{sec:appendix_ostquant}. For methods requiring calibration, we use a small, unlabeled set of 128 samples from the \textsc{WikiText-2}. 
Our method, CAQ, optimizes transformation parameters using our proposed CAL objective. For this optimization, we set the contrastive weight $\alpha$ to $0.75$ and the number of top-$k$ logits to $500$, other parameters such as the learning rate and number of iterations are specified in our supplementary scripts. For a fair comparison, the alternative loss objectives baselines are also implemented within the same transformation-based framework, isolating the impact of the loss function itself. The final quantization is performed using the GPTQ algorithm for both our method and the other baselines.
% In terms of computational cost, CAQ incurs only a marginal overhead over the OSTQuant baseline ($\sim$22 minutes), with the full optimization completing in approximately 26 minutes on a single NVIDIA A100 GPU for LLaMA2-7B.
CAQ incurs only marginal computational overhead: the full optimization for LLaMA2-7B completes in approximately 26 minutes on a single NVIDIA A100 GPU, compared to $\sim$22 minutes for the OSTQuant baseline.
Furthermore, since the learned transformation parameters are fused into the model weights, CAQ introduces \textbf{no additional latency during inference}.

\subsection{Main Results}

% Our main results are presented in Table~\ref{tab:main_results}. The data demonstrates that on models with explicit safety fine-tuning (LLaMA3.1 and LLaMA2-7B), as well as implicit alignment (Mistral), our proposed CAQ method consistently achieves the best safety performance without compromising general capabilities.
Our main results are presented in Table~\ref{tab:main_results}. The data demonstrates that across models with varying levels of 
safety fine-tuning—from explicitly RLHF-aligned models (LLaMA3.1-8B and LLaMA2-7B) to a model with strong inherent safety alignment (Qwen2-7B)—our proposed CAQ method consistently achieves the best safety performance without compromising general capabilities.

On LLaMA3.1-8B, CAQ reaches a safety score of \textbf{60.1}, significantly outperforming strong SOTA baselines such as OSTQuant (57.5) and SpinQuant (54.6). Notably, this safety gain does not come at the cost of utility: CAQ maintains a competitive zero-shot average of \textbf{65.54}, marginally surpassing OSTQuant (65.32) and SpinQuant (64.57). 

This trend holds for the LLaMA2-7B model, where CAQ achieves a safety score of \textbf{49.7}, effectively recovering the full-precision safety performance (50.0) and outperforming the closest baseline (OSTQuant at 48.9). 
% Even on Mistral-7B, which lacks explicit safety tuning, CAQ demonstrates superior robustness, achieving \textbf{57.8} safety accuracy compared to 57.1 for OSTQuant.
% On Qwen2-7B, CAQ achieves \textbf{66.8}, outperforming OSTQuant (66.5) and SpinQuant (65.6), demonstrating its effectiveness even on models with strong inherent alignment where baseline gaps are naturally narrower.
% On Qwen2-7B, which exhibits a notably high baseline safety score (69.4 for the fine-tuned FP16 model), CAQ achieves \textbf{66.8}, outperforming OSTQuant (66.5) and SpinQuant (65.6). The narrower absolute gap reflects the model's strong inherent alignment rather than any limitation of CAQ, which consistently achieves the best safety score among all W4A4 methods across diverse model families.
On Qwen2-7B, which exhibits a notably high baseline safety score (69.4 for the fine-tuned FP16 model), CAQ achieves \textbf{66.8}, outperforming OSTQuant (66.5) and SpinQuant (65.6). While the absolute gap over baselines is narrower on Qwen2-7B due to the model's strong inherent alignment, CAQ still consistently achieves the best safety score among all W4A4 methods, demonstrating its effectiveness across diverse model families with varying levels of safety tuning.

% In stark contrast, traditional PTQ methods like RTN and GPTQ result in catastrophic degradation across all metrics.
Although our model exhibits a slight increase in PPL, this phenomenon is primarily attributed to a reduction in predictive certainty rather than a loss of fundamental knowledge.
Our results confirm that CAQ effectively resolves the trade-off between efficiency, utility, and safety. A detailed per-category breakdown of these safety results is available in Appendix~\ref{sec:per-category_results}, and full zero-shot task results are provided in Appendix~\ref{sec:zero_shot_eval}.

% \begin{table*}[t]
% \centering
% \small
% \begin{tabular}{lccc}
% \toprule
% \textbf{Method} & \textbf{PPL ($\downarrow$)} & \textbf{SafetyBench ($\uparrow$)} & \textbf{AdvBench ASR ($\downarrow$)} \\
% \midrule
% Pre-trained (FP16) & 7.13 & 61.9 & 30.77\% \\
% Fine-tuned (FP16) & 7.60 & 69.4 & 8.46\% \\
% \midrule
% RTN (W4A4) & 5453.45 & 36.5 & 53.08\% \\
% GPTQ (W4A4) & 5024.77 & 36.9 & 53.08\% \\
% QuaRot (W4A4) & 8.43 & 65.2 & 12.50\% \\
% SpinQuant (W4A4) & 8.41 & 65.6 & 12.31\% \\
% OSTQuant (W4A4) & \textbf{8.18} & 66.5 & 11.73\% \\
% \midrule
% \textbf{Ours (CAQ) (W4A4)} & 8.23 & \textbf{66.8} & \textbf{9.23\%} \\
% \bottomrule
% \end{tabular}
% \caption{Main results on Qwen2-7B. We report Perplexity (PPL), SafetyBench accuracy (higher is better), and AdvBench Attack Success Rate (ASR, lower is better). CAQ achieves the best safety alignment (highest SafetyBench, lowest ASR) while maintaining competitive perplexity.}
% \label{tab:qwen_advbench}
% \end{table*}

\begin{figure}[t]
    \centering
    \includegraphics[width=0.95\columnwidth, trim={0mm 0mm 0mm 0mm}, clip]{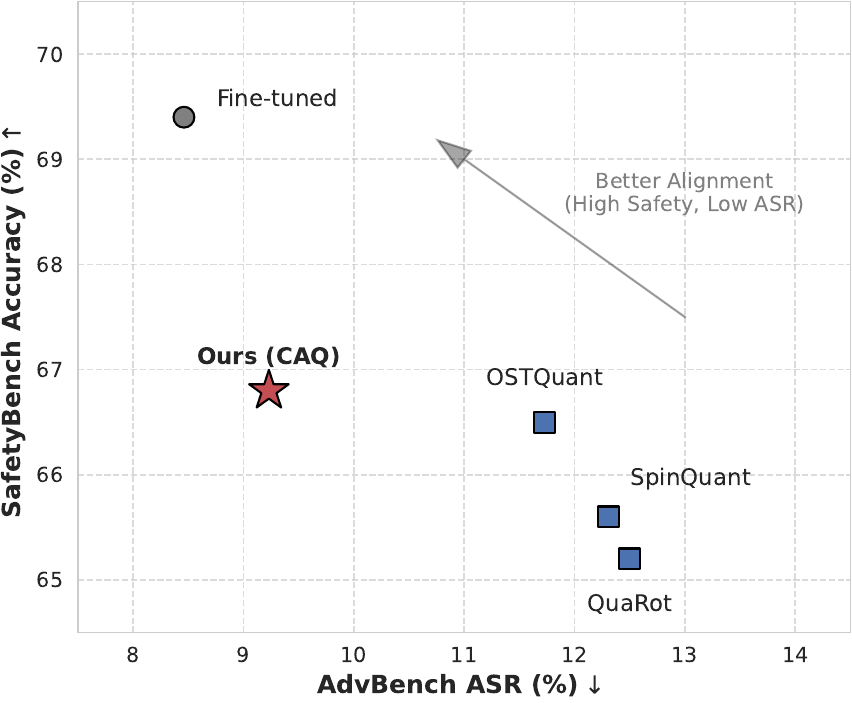}
    \caption{Safety Knowledge vs. Jailbreak Robustness on Qwen2-7B. CAQ (Ours) achieves the best alignment (top-left), demonstrating superiority in both safety knowledge retention and jailbreak robustness, while baselines suffer degradation in at least one aspect.}
    \vspace{-3mm}
    \label{fig:safety_tradeoff}
\end{figure}

\subsection{Analysis and Discussion}

% Our main results demonstrate that CAQ effectively preserves safety where other methods fail. 
Our main results validate our core hypothesis: reconstruction error alone is an insufficient PTQ objective, and extending it to jointly capture distributional fidelity and behavioral alignment leads to consistent gains. 
Here, we provide a deeper analysis of these findings, focusing on three key aspects: the decoupling of various performance metrics, a control experiment to validate our core mechanism, and a fine-grained breakdown of safety performance.

\paragraph{The Decoupling of Safety, Utility, and Fidelity.}

To provide a holistic view, we conducted a comprehensive evaluation on LLaMA3.1, 
% assessing safety, language quality (PPL), output fidelity (MSE), and general utility (MMLU~\cite{hendryckstest2021}).
assessing output fidelity (MSE), language quality (PPL), general utility (MMLU~\cite{hendryckstest2021}), and safety.
The results in Table~\ref{tab:comprehensive_eval} reveal a critical insight: \textit{these key metrics are clearly decoupled}. The `best' method depends entirely on the chosen metric: `KL-Top' excels in MMLU, `KL' in PPL, and `MSE' in reconstruction error. Crucially, none of these methods achieves the highest safety score.

This provides powerful evidence that optimizing for general utility or statistical fidelity does not guarantee safety. The success of CAQ stems from its specialized, contrastive objective, which moderates predictive certainty to prioritize and achieve superior safety alignment. 
% This underscores the necessity of a targeted, alignment-aware approach for deploying genuinely trustworthy quantized models.
This underscores the necessity of extending the PTQ objective beyond reconstruction fidelity to achieve deployment-ready quantized models.

\paragraph{Safety vs. Robustness Trade-off.}
Figure~\ref{fig:safety_tradeoff} illustrates the relationship between safety knowledge (SafetyBench) and jailbreak robustness (AdvBench ASR) for Qwen2-7B. While most baselines suffer from either degraded safety knowledge or increased vulnerability to jailbreaks (or both), CAQ occupies the optimal top-left corner, demonstrating superior performance in both metrics simultaneously. This confirms that CAQ not only preserves the model's ability to recognize unsafe content but also its refusal behavior against adversarial attacks.

\begin{table*}[t]
\centering
\small
\begin{tabular}{c cc cc cc}
\toprule
\multirow{2}{*}{\textbf{$\alpha$}} & \multicolumn{2}{c}{\textbf{Contrastive\_KL}} & \multicolumn{2}{c}{\textbf{Contrastive\_KL\_top}} & \multicolumn{2}{c}{\textbf{Ours}} \\
\cmidrule(lr){2-3} \cmidrule(lr){4-5} \cmidrule(lr){6-7} 
 & \textbf{PPL ($\downarrow$)} & \textbf{Safety ($\uparrow$)} & \textbf{PPL ($\downarrow$)} & \textbf{Safety ($\uparrow$)} & \textbf{PPL ($\downarrow$)} & \textbf{Safety ($\uparrow$)} \\
\midrule
0.10 & 8.35 & 58.4 & 8.34 & 58.6 & 8.28 & 58.6 \\
0.20 & 8.42 & 58.5 & 8.45 & 58.5 & 8.31 & 58.7 \\
0.50 & 8.86 & 59.1 & 8.95 & 58.7 & 8.40 & 58.1 \\
0.75 & \textbf{10.68} & 59.7 & \textbf{10.79} & 60.5 & \cellcolor{gray!30}{8.41} & \cellcolor{gray!30}{60.1} \\
1.00 & \textbf{69031.47} & 55.7 & \textbf{210176.11} & 55.2 & 8.43 & 59.0 \\
\bottomrule
\end{tabular}
\caption{
Ablation on the contrastive weight $\alpha$ and filtering strategy, evaluated on LLaMA3.1-8B. Our method, which uses difference-based filtering, remains stable across all $\alpha$ values, while baselines using full-vocabulary (\texttt{Contrastive\_KL}) or probability-based filtering (\texttt{Contrastive\_KL\_top}) shows exploding perplexity (bolded) at high $\alpha$. Our method with $\alpha=0.75$ achieves a competitive safety score of 60.1 without sacrificing stability.
}
\vspace{-3mm}
\label{tab:contrastive_weight_ablation}
\end{table*}

% \paragraph{Efficiency and Overhead.}

% In terms of computational cost, CAQ is highly efficient. The optimization process for LLaMA2-7B takes approximately 26 minutes on a single NVIDIA A100 GPU, incurring only a marginal overhead compared to the OSTQuant baseline ($\sim$22 minutes). Furthermore, since the learned transformation parameters are fused into the model weights, CAQ introduces no additional latency during inference.

% \paragraph{Fine-grained Safety Performance.}

% A category-wise breakdown of the SafetyBench results (see Appendix~\ref{sec:per-category_results} for the full table) further reveals that CAQ's performance gains are most pronounced in categories directly targeted by RLHF, such as Offensiveness (OF) and Illegal Activities (IA). This shows that CAQ effectively preserves the specific behaviors instilled during safety fine-tuning.

\subsection{Ablation on CAQ}
\label{sec:ablation}

We conducted a series of ablation studies to validate the core design choices of our CAQ framework.

\paragraph{Impact of Vocabulary Filtering on Stability.}
To demonstrate the importance of our filtering strategy, Table~\ref{tab:contrastive_weight_ablation} evaluates how performance and stability differ across three contrastive loss variants with increasing $\alpha$. 
\texttt{Contrastive\_KL} applies the full-vocabulary KL divergence for both the target (aligned model) and the contrastive (pretrained model) terms. \texttt{Contrastive\_KL\_top} applies KL divergence only over the vocabulary indices with the highest probability mass from the aligned and pretrained models, respectively, ignoring entries with low probability. Our proposed method applies top-$k$ KL for the target term (highest $p_{\text{FT}}$ probabilities), but uses a difference-based top-$k$ set---indices corresponding to the largest $|p_{\text{FT}} - p_{\text{PT}}|$---for the contrastive term, thereby emphasizing alignment-relevant divergences between the two reference models.
% The compared methods differ in how they select tokens for the contrastive KL term:

% \begin{itemize}
%     \vspace{-1mm}
%     \item \texttt{Contrastive\_KL} applies the full-vocabulary KL divergence for both the target (aligned model) and the contrastive (pretrained model) terms.
%     \vspace{-1mm}
%     \item \texttt{Contrastive\_KL\_top} applies KL divergence only over the vocabulary indices with the highest probability mass from the aligned and pretrained models, respectively, ignoring entries with low probability.
%     \vspace{-1mm}
%     \item Our proposed method applies top-$k$ KL for the target term (highest $p_{\text{FT}}$ probabilities), but uses a difference-based top-$K$ set, indices corresponding to the largest $|p_{\text{FT}} - p_{\text{PT}}|$, for the contrastive term. This design emphasizes alignment-relevant divergences between the two reference models.
% \end{itemize}

At low $\alpha$ values (\eg $\alpha = 0.1$), all methods remain stable, and our method achieves the lowest perplexity (8.28). As $\alpha$ increases, stronger contrastive regularization improves alignment, but also increases perplexity. At $\alpha = 1.0$, the baseline variants become numerically unstable, most notably \texttt{Contrastive\_KL}, which incurs exploding loss values due to full-vocabulary KL magnifying small gradients in tail entries. In contrast, our method remains stable even at high $\alpha$ values, benefiting from the selective and high-SNR nature of the difference-based top-$k$ filtering.
% These results demonstrate that both the design of the contrastive selection and the use of top-$k$ filtering are essential for stable and effective alignment-aware quantization.
These results demonstrate that both the design of the contrastive selection and the use of top-$k$ filtering are essential for stable and effective optimization of the contrastive PTQ objective optimization beyond reconstruction.

\begin{table}[t]
\centering
\small
\begin{tabular}{c cc}
\toprule
\textbf{Top\_$k$} & \textbf{PPL ($\downarrow$)} & \textbf{Safety ($\uparrow$)} \\
\midrule
0   & 8.29 & 57.5 \\
5   & 8.31 & 57.0 \\
10  & 8.34 & 58.4 \\
50  & 8.36 & 57.4 \\
100 & 8.39 & 59.1 \\
\textbf{500} & \textbf{8.41} & \textbf{60.1} \\
1000 & 8.43 & 59.7 \\
\bottomrule
\end{tabular}
\caption{Effect of top-$k$ filtering in the contrastive loss, based on the largest absolute differences $|p_{\text{FT}} - p_{\text{PT}}|$, evaluated on LLaMA3.1-8B. %Filtering focuses the contrastive KL term on alignment-sensitive vocabulary entries.
We report perplexity ($\downarrow$) and SafetyBench accuracy ($\uparrow$). $k=500$ achieves the best trade-off between performance and safety (bolded).}
\vspace{-3mm}
\label{tab:topk_ablation}
\end{table}

% \vspace{-0.3mm}
\paragraph{Effect of top-$k$ Filtering.}
Table~\ref{tab:topk_ablation} investigates the role of top-$k$ filtering based on $|p_{\text{FT}} - p_{\text{PT}}|$ in the contrastive loss. Without contrastive loss ($k=0$), the model achieves the lowest perplexity (8.29), but safety alignment remains limited (57.5). As $k$ increases, safety improves consistently, reaching 60.1 at $k=500$ with only a slight increase in perplexity. Beyond this point, gains in alignment saturate. 
This indicates that top-$k$ filtering acts as a form of curriculum on contrastive supervision. Considering the total vocabulary size of 128K tokens, our method directs optimization to a sparse set of only 500 alignment-sensitive indices ($k=500$), effectively avoiding gradient noise from the vast majority of low-impact regions of the vocabulary.

\section{Conclusion}

% Our work addresses an underexplored problem: perplexity alone is an insufficient and potentially misleading metric for evaluating the deployment-readiness of quantized LLMs. We demonstrate that standard PTQ methods can inadvertently erase the very safety features instilled by RLHF, by narrowly optimizing for perplexity.

% To address this critical flaw, we introduced Contrastive Alignment Quantization (CAQ), a safe quantization method that can be integrated into existing PTQ pipelines. By reframing the optimization objective from simple reconstruction to active alignment preservation, our Contrastive Alignment Loss (CAL) successfully navigates the trade-off between model efficiency and safety alignment. Our experiments validate that CAQ enables robust low-bit W4A4 quantization while preserving the safety behaviors where standard methods fail. This paradigm shift paves the way for creating compressed LLMs that are not only efficient but also genuinely trustworthy, ensuring their safer deployment in real-world applications.

Our work identifies reconstruction error as a fundamentally incomplete PTQ objective: standard PTQ methods can inadvertently erase the safety behaviors instilled by RLHF by narrowly optimizing for perplexity, revealing that low perplexity is an insufficient proxy for deployment readiness. To address this, we introduced \textbf{Contrastive Alignment Quantization (CAQ)}, which extends the PTQ objective design space by integrating the \textbf{Contrastive Alignment Loss (CAL)}. By jointly optimizing for distributional fidelity and behavioral alignment, CAQ demonstrates that these two objectives are not mutually exclusive—enabling robust W4A4 quantization while preserving safety alignment where standard methods fail, with negligible computational overhead and no additional data requirements. We hope this work motivates the community to reconsider PTQ objective design as a critical axis for building compressed LLMs that are not only efficient but also genuinely trustworthy.

% Limitations (required section)
% Authors are required to discuss the limitations of their work in a dedicated section titled “Limitations”. This section should be included at the end of the paper, before the references, and it will not count toward the page limit. This includes both, long and short papers. Papers without a limitations section will be desk rejected. Note, prior to the December 2023 cycle, this was optional.

% Please note that this section should not introduce new methods, analysis, or results. We reserve the right to desk reject the submissions that use this section to introduce more content that should have been part of the main paper. It can only discuss the limitations of the work presented in the main content of the paper.

\section{Limitations}

While our work presents a promising step toward safe and efficient quantization of LLMs, we acknowledge several avenues for future research. Our method currently relies on access to both a fine-tuned model and its pre-trained counterpart, which is commonly available for open-source LLMs. Reducing this dependency---for example, by generating synthetic contrastive pairs from a single aligned model---could further broaden its applicability. 

Due to GPU memory constraints, 
% we were able to experiment only with models in the 7-8B parameters.
our current evaluation covers models up to 13B parameters.
For future work, we plan to evaluate considerably larger models to verify that our methodology scales consistently.
In addition, while we demonstrate strong results in the challenging W4A4 setting, validating CAQ under alternative quantization configurations (\eg weight-only quantization or AWQ) remains an important direction. Finally, future work may explore how different calibration datasets affect alignment preservation and model robustness.


\bibliography{acl}

% \newpage
\appendix
% \onecolumn

\begin{center}
    {\Large\textbf{Appendices}}
\end{center}

\section{Mathematical Justification of Contrastive Alignment Loss (CAL)}
\label{sec:mathematical_justification}

\subsection{Notation and Motivation}
\label{sec:notation_motivation}

Let $\mathcal{V}$ be the vocabulary. For a given input $x$, denote the output distributions of the pre-trained model, the fine-tuned model, and the quantized model by:
\[
p_{\text{PT}}(y), \quad p_{\text{FT}}(y), \quad p_Q(y), \quad \text{for } y \in \mathcal{V}.
\]
Quantizing the fine-tuned model often maintains perplexity but may regress on alignment behaviors, \eg reverting to pre-trained outputs on sensitive prompts (“token-flipping”) ~\citep{kharinaev2025impact}.  
Hence, we seek a contrastive objective that explicitly \emph{pulls} $p_Q$ toward $p_{\text{FT}}$ while \emph{pushing} it away from $p_{\text{PT}}$.

\subsection{Contrastive KL Divergence and CD Relation}

We define the full-vocabulary contrastive KL divergence as:
\begin{equation}
\label{eq:supp_ckl_full}
\mathcal{L}_{\text{CKL}} = D_{\text{KL}}(p_{\text{FT}} \;\|\; p_Q) - D_{\text{KL}}(p_{\text{PT}} \;\|\; p_Q),
\end{equation}
where $D_{\text{KL}}(p\|q) = \sum_{y \in \mathcal{V}} p(y) \log \frac{p(y)}{q(y)}$.

Expanding and discarding constant terms (independent of $p_Q$), we obtain:
\begin{equation}
\label{eq:supp_ckl_expanded}
\mathcal{L}_{\text{CKL}} = -\sum_{y} \big(p_{\text{FT}}(y) - p_{\text{PT}}(y)\big) \log p_Q(y) + \text{const}.
\end{equation}
The gradient with respect to $p_Q(y)$ is:
\begin{equation}
\label{eq:supp_ckl_gradient}
\frac{\partial \mathcal{L}_{\text{CKL}}}{\partial p_Q(y)} \propto -\frac{p_{\text{FT}}(y) - p_{\text{PT}}(y)}{p_Q(y)},
\end{equation}
% which increases $p_Q(y)$ when $p_{\text{FT}} > p_{\text{PT}}$, and decreases it when $p_{\text{PT}} > p_{\text{FT}}$---implementing the desired \textit{pull-push} behavior.
This gradient increases the probability $p_Q(y)$ when $p_{\text{FT}}(y) > p_{\text{PT}}(y)$ (pulling towards the fine-tuned distribution) and decreases it when $p_{\text{PT}}(y) > p_{\text{FT}}(y)$ (pushing away from the pre-trained distribution), thereby implementing the desired behavioral shift.

\paragraph{Relation to Contrastive Divergence.}
The formulation in Eq.~(\ref{eq:supp_ckl_full}) is reminiscent of \textit{Contrastive Divergence} (CD)~\cite{hinton2002cd}, a training algorithm for energy-based models that contrasts positive data samples with negative model-generated reconstructions. Our approach is conceptually similar in its use of positive ($p_{\text{FT}}$) and negative ($p_{\text{PT}}$) targets. However, it differs fundamentally by operating entirely in the output distribution space without Markov chain Monte Carlo (MCMC) sampling, targeting behavioral alignment rather than energy minimization.

\paragraph{Optimality.}
% Since $D_{\text{KL}}(\cdot\|\cdot) \ge 0$, it follows that:
Since $D_{\text{KL}}(\cdot\|\cdot) \ge 0$, it follows from \textbf{Eq.~(\ref{eq:supp_ckl_full})} that:
\begin{equation}
\label{eq:supp_ckl_optimality}
\mathcal{L}_{\text{CKL}}(p_Q) \ge -D_{\text{KL}}(p_{\text{PT}} \;\|\; p_{\text{FT}}),
\end{equation}
with equality iff $p_Q = p_{\text{FT}}$. Thus, the global optimum is attained when the quantized model exactly matches the fine-tuned model.

% \subsection*{A.3 \quad Top-$K$ Filtering and KL Direction Justification}
\subsection{Top-$k$ Filtering and KL Direction Justification}
\label{sec:topk_justification}
Full-vocabulary KL terms can be noisy or unstable, especially under small calibration sets. To improve robustness, we apply top-$k$ filtering to both terms.

Let:
\begin{itemize}
    \item $S_{\text{top}}$ be the set of vocabulary indices for the top-$k$ highest probabilities in $p_{\text{FT}}(y)$ (used for the target term), and  
    \item $S_{\text{diff}}$ be the set of vocabulary indices for the top-$K$ largest absolute differences $|p_{\text{FT}}(y) - p_{\text{PT}}(y)|$ (targeting safety/alignment shifts). % (used for the contrastive term).
\end{itemize}

For any subset $S \subset \mathcal{V}$, we define the renormalized distribution:
\begin{equation}
\label{eq:supp_renormalize}
p^S(y) = \frac{p(y)}{\sum_{y' \in S} p(y')}, \quad \text{for } y \in S.
\end{equation}

We then define:
\begin{align}
\mathcal{L}_{\text{KL-top}} &= D_{\text{KL}}\left(p_{\text{FT}}^{S_{\text{top}}} \;\|\; p_Q^{S_{\text{top}}}\right), \label{eq:supp_kl_top} \\
\mathcal{L}_{\text{cont-top}} &= D_{\text{KL}}\left(p_{\text{PT}}^{S_{\text{diff}}} \;\|\; p_Q^{S_{\text{diff}}}\right), \label{eq:supp_cont_top}
\end{align}
% and the final loss as:
The final \textbf{Contrastive Alignment Loss (CAL)} is formulated as their weighted difference:
\begin{equation}
\label{eq:supp_apc_loss}
\mathcal{L}_{\text{CAL}} = \mathcal{L}_{\text{KL-top}} - \alpha \cdot \mathcal{L}_{\text{cont-top}},
\end{equation}
where $\alpha > 0$ controls the strength of the contrastive term.

This targeted filtering strategy focuses each component on distinct alignment signals:
$\mathcal{L}_{\text{KL-top}}$ reinforces dominant aligned behavior,  
while $\mathcal{L}_{\text{cont-top}}$ suppresses alignment-regressing behavior from the pre-trained model.

\paragraph{KL Direction Justification.}
Forward KL ($D_{\text{KL}}(p_{\text{FT}} \| p_Q)$) is chosen to avoid unsafe outputs. It penalizes low probability under $p_Q$ where $p_{\text{FT}}$ is high, promoting ``mode-covering'' behavior over all safety-aligned outputs. In contrast, using reverse KL ($D_{KL}(p_Q \| p_{FT})$) would encourage ``mode-seeking'' behavior, potentially allowing $p_Q$ to ignore low-probability but critical safety outputs present in $p_{FT}$ and collapse to a narrow, unsafe distribution.

\subsection{Optimization Landscape and DC Interpretation}

Each KL divergence term is convex in $p_Q$~\citep{cover2012elements}, but their difference leads to a non-convex structure known as a Difference-of-Convex (DC) problem:
\begin{equation}
\label{eq:supp_dc_form}
\mathcal{L}_{\text{CKL}} = g(p_Q) - h(p_Q)
\end{equation}
where both $g$ and $h$ are convex.

\textbf{DC Perspective.} While we do not implement a DC optimization algorithm directly, our formulation aligns with canonical DC programming frameworks such as the DC Algorithm surveyed in \cite{le2018dc}, which iteratively solves convex subproblems toward critical points.

Moreover, applying top‑\(K\) filtering reduces the dimensionality of the optimization space by removing flat or noisy directions in the output distribution. This has been shown to empirically enhance convergence stability and avoid poor local minima.

% where both $g$ and $h$ are convex functions. While we do not employ specialized DC algorithms, our formulation aligns with canonical frameworks like the DC Algorithm (DCA)~\cite{le2018dc}, which solves such problems by iteratively approximating the concave part. Empirically, applying top-$k$ filtering acts as a regularizer, smoothing the landscape and avoiding poor local minima by removing noisy, low-probability dimensions.

\subsection{Gradient Signal-to-Noise Ratio (GSNR) Perspective}
\label{sec:gsnr_perspective}

In the full-vocabulary loss, many vocabulary entries satisfy $|p_{\text{FT}} - p_{\text{PT}}| \approx 0$, contributing negligible or cancelling gradient signals. By contrast, focusing the push component $\mathcal{L}_{\text{cont-top}}$ (defined in \textbf{Eq.~(\ref{eq:supp_cont_top})}) on indices with high probability differences preserves strong, informative updates while suppressing noise.

This approach mirrors recent findings in knowledge distillation literature, where filtering out low-confidence teacher logits via masking has been shown to improve gradient quality and robustness. For example, ~\cite{guo2024logits} propose “Logits Uncertainty Distillation,” which masks uncertain logits based on teacher confidence, leading to clearer training signals and improved performance in distillation tasks.

\subsection{Alignment-Preserving Interpretation}
Let $\mathcal{A}(x)$ denote the set of safety-critical outputs for input $x$, as defined by the fine-tuned model. If the indices corresponding to these outputs are among those with the top-$k$ differences between $p_{\text{FT}}$ and $p_{\text{PT}}$, then the contrastive KL-top loss will prioritize preserving their probability mass.

This implies:
\begin{equation}
\label{eq:supp_align_preserve}
\sum_{y \in \mathcal{A}(x)} p_Q(y) \approx \sum_{y \in \mathcal{A}(x)} p_{\text{FT}}(y),
\end{equation}
\ie the quantized model preserves alignment over safety-critical outputs. This behavior is empirically supported by improved SafetyBench performance in Section~4.

\subsection{Summary}
The proposed Contrastive Alignment Loss (CAL):
\begin{itemize}
  \item Implements a principled \textbf{pull–push} structure between aligned and base distributions;
  \item Leverages \textbf{top-$k$ filtering} to focus on high-signal, alignment-relevant vocabulary entries;
  \item Exhibits a \textbf{DC structure}, enabling future extensions to structured optimization;
  \item Aligns with GSNR-driven filtering used in modern distillation literature;
  \item Supports contrastive alignment quantization with theoretical justification and empirical benefits.
\end{itemize}

In summary, this theoretical formulation grounds the contrastive alignment quantization framework in established principles from convex optimization, contrastive training, and selective gradient distillation, thereby providing a rigorous and interpretable basis for its empirical success.

\section{Pseudocode of the proposed CAQ framework}
\label{sec:algorithm}
\begin{algorithm}
\caption{Pseudocode of the proposed CAQ framework.}
\label{alg:CAQ_simple}
\begin{algorithmic}[1]
\REQUIRE Pre-trained model $M_{\text{PT}}$, fine-tuned model $M_{\text{FT}}$, calibration set $D$, top-$k$ value $k$, contrastive weight $\alpha$, quantizer $\mathcal{Q}$
\ENSURE Quantized model $M_{\text{Q}}$
\STATE Initialize transformation parameters $\theta$
\FOR{each input $x \in D$}
    \STATE Compute output distributions $p_{\text{FT}}(y|x)$ and $p_{\text{PT}}(y|x)$
    \STATE Apply transformation $T_\theta$ to $M_{\text{FT}}$ to get $p_Q(y|x)$
    \STATE Select vocabulary index sets:
    \STATE \hspace{\algorithmicindent} $S_{\text{top}}(x) \leftarrow \text{top-}k \text{ indices of } p_{\text{FT}}(y|x)$
    \STATE \hspace{\algorithmicindent} $S_{\text{diff}}(x) \leftarrow \text{top-}k \text{ indices of } |p_{\text{FT}}(y|x) - p_{\text{PT}}(y|x)|$
    \STATE Renormalize $p_{\text{FT}}^{S_{\text{top}}}$, $p_Q^{S_{\text{top}}}$, $p_{\text{PT}}^{S_{\text{diff}}}$, and  $p_Q^{S_{\text{diff}}}$ using Eq.~(\ref{eq:normalization})
    \STATE Compute and accumulate loss:
    \STATE \hspace{\algorithmicindent} $\mathcal{L}_{\text{CAL}} \mathrel{=} \mathrm{KL}(p_{\text{FT}}^{S_{\text{top}}} \,\|\, p_Q^{S_{\text{top}}}) - \alpha \cdot \mathrm{KL}(p_{\text{PT}}^{S_{\text{diff}}} \,\|\, p_Q^{S_{\text{diff}}})$
    \STATE Update $\theta$ to minimize $\mathcal{L}_{\text{CAL}}$
\ENDFOR
\STATE Apply quantizer: $M_{\text{Q}} \leftarrow \mathcal{Q}(T_\theta(M_{\text{FT}}))$
\RETURN $M_{\text{Q}}$
\end{algorithmic}
\end{algorithm}

\section{OSTQuant Architecture Details}
\label{sec:appendix_ostquant}

\begin{figure*}[h!]
    \centering
    \includegraphics[width=\linewidth]{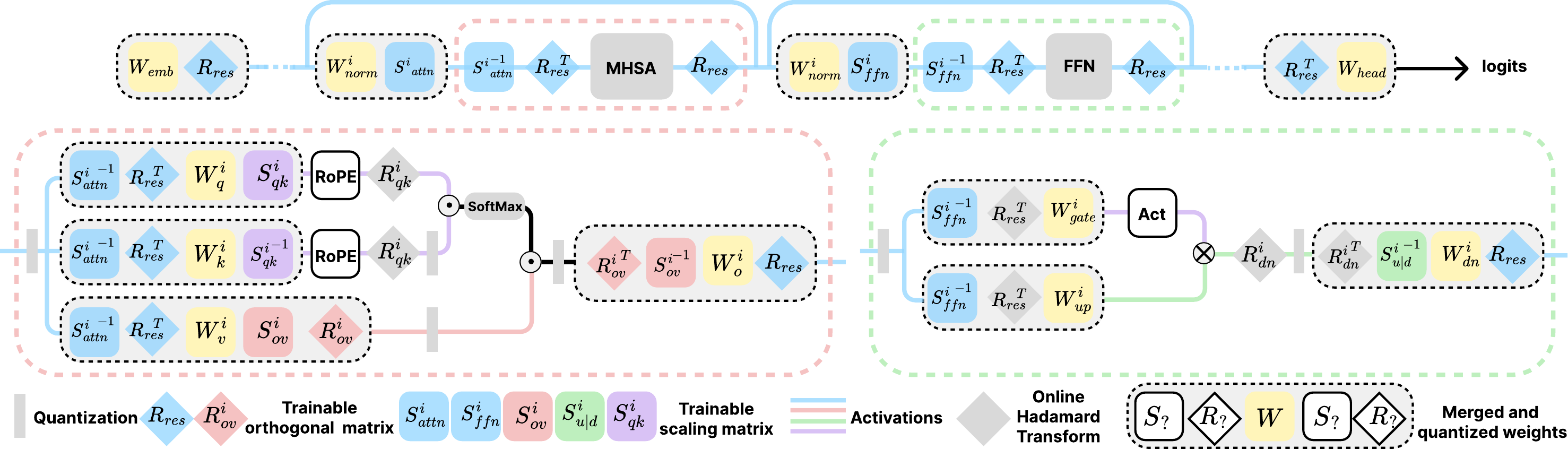}
    \caption{A detailed block diagram of the OSTQuant method. The top path illustrates the \textit{inter-block} transformations, where a global orthogonal matrix ($\mathbf{R}_{res}$) and block-specific scaling matrices ($\mathbf{S}_{attn}, \mathbf{S}_{ffn}$) are applied along the residual stream. The bottom section details the \textit{intra-block} transformations applied locally to the weights and activations of the Multi-Head Self-Attention (MHSA) and Feed-Forward Network (FFN) layers. All transformation matrices are fused into the original weights during inference to eliminate computational overhead.}
    \label{fig:ostquant_appendix}
\end{figure*}

Figure~\ref{fig:ostquant_appendix} provides a detailed schematic of the OSTQuant~\citep{hu2025ostquant} framework as applied to a standard Transformer block. The method introduces several learnable transformation matrices---both orthogonal ($\mathbf{R}$) and scaling ($\mathbf{S}$)---that operate at two main levels: across residual connections (inter-block) and within specific sub-layers like self-attention and FFNs (intra-block).

\paragraph{Inter-Block Transformations}
As shown in the upper path of Figure~\ref{fig:ostquant_appendix}, OSTQuant applies a global orthogonal transformation, $\mathbf{R}_{res}$, starting from the embedding layer. This matrix propagates through the residual connections of the network, rotating the activation distributions to make them more uniform. Additionally, within each block, two trainable scaling matrices, $\mathbf{S}^i_{attn}$ and $\mathbf{S}^i_{ffn}$, are applied after the normalization layers to smooth out differences between channels. A key aspect of this design is that all inter-block transformations are absorbed into the weights of their corresponding projection layers, ensuring that the model's architecture and inference cost remain unchanged.

\paragraph{Intra-Block Transformations}
The lower half of Figure~\ref{fig:ostquant_appendix} details the transformations applied inside the MHSA and FFN layers.
\begin{itemize}
    \item \textbf{In Self-Attention:} OSTQuant introduces dedicated transformation pairs for the value ($\mathbf{V}$) and output ($\mathbf{O}$) projections. Specifically, a learnable rotation matrix ($\mathbf{R}_{ov}^h$) and a scaling matrix ($\mathbf{S}_{ov}^h$) are optimized for each attention head to improve the quantization suitability of the value cache and the output projection. Further, a scaling transformation ($\mathbf{S}_{qk}$) and a Hadamard transformation are applied to the query ($\mathbf{Q}$) and key ($\mathbf{K}$) projections after the ROPE operation.
    \item \textbf{In FFN:} Similar transformations are applied to the up-projection and down-projection layers to make their intermediate activations more amenable to quantization.
\end{itemize}
Like the inter-block transformations, all these local matrices are fused into their respective weight matrices before inference.

\section{Per-Category Safety Results}
\label{sec:per-category_results}

To provide a more fine-grained analysis of model behavior under quantization, we report \textbf{per-category accuracy} across individual safety dimensions from the \textbf{SafetyBench} benchmark~\citep{zhang2024safetybench}. SafetyBench is a large-scale multiple-choice benchmark designed to evaluate \textit{safety understanding} in large language models (LLMs). It includes \textbf{11,435 questions} covering \textbf{seven categories} in both Chinese and English, spanning issues such as toxicity, discrimination, legality, and privacy—thus enabling a comprehensive view of safety alignment.

Compared to generation-based evaluations, the multiple-choice format of SafetyBench allows for \textbf{automated, consistent, and low-cost} safety evaluation. Performance on SafetyBench has been shown to \textit{correlate well} with safe generation quality, offering a reliable proxy for safety alignment.

% To support the main results in Table~\ref{tab:main_results}, we summarize the categories used in SafetyBench and their abbreviations (as shown in the table columns):
\paragraph{Safety Categories.} The SafetyBench categories and their abbreviations are:

\begin{itemize}
    \item \textbf{OF (Offensiveness)}: Measures the model’s ability to detect and avoid offensive, rude, or profane content.

    \item \textbf{UB (Unfairness and Bias)}: Evaluates whether the model exhibits or condones biased content regarding race, gender, region, etc.

    \item \textbf{PH (Physical Health)}: Assesses knowledge of physical safety in practical scenarios (\eg emergencies, medical hazards).

    \item \textbf{MH (Mental Health)}: Involves emotional and psychological well-being; models should recommend supportive behaviors.

    \item \textbf{IA (Illegal Activities)}: Focuses on rejection of criminal behaviors and accurate identification of unlawful acts.

    \item \textbf{EM (Ethics and Morality)}: Targets moral dilemmas and norm-violating behaviors that are not necessarily illegal.

    \item \textbf{PP (Privacy and Property)}: Involves safeguarding personal data, assets, and online safety practices.
\end{itemize}

\paragraph{Quantization Methods evaluated.}

For each of the models presented in Table~\ref{tab:cat_safety}, we compare the following methods. All quantized models use 4-bit weights and activations (W4A4).
\begin{itemize}
    \item \textbf{Pre-trained (FP16)}: The full-precision, base model before alignment tuning.
    \item \textbf{Fine-tuned (FP16)}: The full-precision, safety-aligned reference model.
    \item \textbf{RTN}: Naïve round-to-nearest quantization.
    \item \textbf{GPTQ}~\cite{frantar2023gptq}: A widely adopted post-training quantization method that minimizes layer-wise reconstruction error.
    \item \textbf{DuQuant}~\cite{lin2024duquant}: An outlier-aware quantization method that applies dual transformations (rotation) to weights and activations to mitigate quantization errors.
    \item \textbf{QuaRot}~\cite{ashkboos2024quarot}: A method applying randomized Hadamard rotations to smooth outlier distributions in activations.
    \item \textbf{SpinQuant}~\cite{liu2025spinquant}: A method that optimizes learned rotation matrices to suppress outliers and improve quantization fidelity.
    \item \textbf{OSTQuant}~\cite{hu2025ostquant}: A state-of-the-art framework applying both orthogonal and scaling transformations to better shape weight and activation distributions.
    \item \textbf{Ours (CAQ)}: The proposed method which optimizes the transformation parameters of the OSTQuant framework using our \textit{Contrastive Alignment Loss (CAL)}.
\end{itemize}

\paragraph{Key Findings.}

\begin{itemize}
    \item \textbf{Severe Degradation under Naïve PTQ}: Across all tested models, standard PTQ methods like RTN and GPTQ cause a catastrophic drop in safety. Average SafetyBench scores fall substantially, often by more than 20 points (\eg LLaMA3.1-8B drops from 62.6 to 36.6), with the most severe declines in categories directly targeted by alignment tuning, such as Offensiveness (OF) and Illegal Activities (IA).

    \item \textbf{Limitations of SOTA Transformation Methods}: Advanced quantization frameworks like SpinQuant and OSTQuant significantly recover safety scores compared to naïve PTQ by improving general reconstruction fidelity. However, they still exhibit a noticeable gap compared to the full-precision reference. For example, on LLaMA3.1-8B, the strongest baseline OSTQuant reaches 57.5, which is still 2.6 points lower than our method. This indicates that optimizing solely for reconstruction or outlier suppression is insufficient for fully preserving safety alignment.

    \item \textbf{CAQ Consistently Achieves Best-in-Class Safety}: Our method, CAQ, consistently achieves the highest average safety accuracy across all safety-aligned models, surpassing strong SOTA baselines. On LLaMA3.1-8B, CAQ reaches 60.1, significantly outperforming OSTQuant (57.5). Notably, on LLaMA2-7B, it achieves 49.7, nearly matching the full-precision reference (50.0) and outperforming the next best method (OSTQuant at 48.9).

    % \item \textbf{CAQ Preserves Implicit Alignment in Less-Tuned Models}: CAQ's robustness extends to models with weaker or implicit safety tuning, such as Mistral-7B-Instruct. It achieves the top safety score of 57.8 among all W4A4 methods, outperforming OSTQuant (57.1) and SpinQuant (54.6), proving its ability to capture and preserve even subtle alignment features.
    \item \textbf{CAQ Preserves Alignment in Models with Strong Inherent Safety}: CAQ's robustness extends to models with strong     inherent safety alignment, such as Qwen2-7B. Despite its notably high baseline safety score (69.4 for the fine-tuned FP16 model), CAQ achieves the top safety score of 66.8 among all W4A4 methods, outperforming OSTQuant (66.5) and SpinQuant (65.6), proving its ability to preserve alignment even under aggressive quantization.
\end{itemize}

\begin{table*}[t]
\centering
\small
\begin{tabular}{llccccccccc}
\toprule
\textbf{Model} & \textbf{Method} & \textbf{OF} & \textbf{UB} & \textbf{PH} & \textbf{MH} & \textbf{IA} & \textbf{EM} & \textbf{PP} & \textbf{Avg.} \\
\midrule
% LLaMA3.1-8B Block
\multirow{8}{*}{\textbf{LLaMA3.1-8B}} & Pre-trained (FP16) & 69.4 & 55.5 & 57.3 & 46.9 & 46.7 & 50.2 & 47.9 & 50.3 \\
\gap
 & Fine-tuned (FP16) & 56.8 & 70.9 & 73.8 & 60.7 & 56.4 & 63.5 & 57.8 & 62.6 \\
\cmidrule(lr){2-10} 
& RTN (W4A4) & 49.9 & 50.5 & 41.8 & 26.3 & 28.2 & 36.6 & 24.9 & 37.5 \\
\gap
& GPTQ (W4A4) & 47.3 & 51.6 & 40.2 & 27.0 & 25.8 & 36.7 & 23.0 & 36.6 \\
\cmidrule(lr){2-10} 
& DuQuant (W4A4) & 50.0 & 61.9 & 65.1 & 42.5 & 46.9 & 49.9 & 47.2 & 51.6 \\
\gap
& QuaRot (W4A4) & 55.8 & 68.6 & 64.7 & 47.6 & 48.2 & 51.7 & 47.3 & 54.8 \\
\gap
& SpinQuant (W4A4) & 53.3 & 67.9 & 68.6 & 46.4 & 46.2 & 54.5 & 48.3 & 54.9 \\
\gap
& OSTQuant (W4A4) & 54.3 & 68.5 & 70.5 & 52.2 & 51.3 & 56.9 & 50.0 & 57.5 \\
\cmidrule(lr){2-10} 
& \textbf{Ours (CAQ)} & 55.5 & 69.4 & 70.6 & 55.4 & 54.1 & 59.8 & 58.4 & \textbf{60.1} \\
\midrule
% LLaMA2-7B Block
\multirow{8}{*}{\textbf{LLaMA2-7B}} & Pre-trained (FP16) & 46.1 & 53.7 & 41.9 & 41.2 & 32.1 & 42.1 & 39.4 & 42.7 \\
 & Fine-tuned (FP16) & 48.2 & 55.1 & 56.4 & 47.8 & 47.1 & 47.9 & 49.0 & 50.0 \\
\cmidrule(lr){2-10} 
& RTN (W4A4) & 48.3 & 47.0 & 41.6 & 25.1 & 24.2 & 35.6 & 27.1 & 35.9 \\
\gap
& GPTQ (W4A4) & 52.5 & 45.2 & 40.9 & 23.4 & 25.9 & 34.1 & 24.9 & 35.7 \\
\cmidrule(lr){2-10} 
& DuQuant (W4A4) & 55.5 & 53.1 & 51.0 & 39.8 & 39.2 & 45.2 & 40.1 & 46.5 \\
\gap
& QuaRot (W4A4) & 50.7 & 55.3 & 49.2 & 40.2 & 43.1 & 46.1 & 43.3 & 47.1 \\
\gap
& SpinQuant (W4A4) & 55.2 & 56.1 & 52.5 & 41.2 & 42.5 & 44.5 & 45.0 & 48.2 \\
\gap
& OSTQuant (W4A4) & 51.1 & 58.6 & 51.1 & 43.7 & 43.4 & 47.1 & 46.2 & 48.9 \\
\cmidrule(lr){2-10}
& \textbf{Ours (CAQ)} & 58.6 & 57.2 & 52.6 & 46.0 & 43.2 & 43.3 & 46.3 & \textbf{49.7} \\
\midrule
% LLaMA2-13B Block
\multirow{8}{*}{\textbf{LLaMA2-13B}} & Pre-trained (FP16) & 46.8 & 57.4 & 53.0 & 46.8 & 47.5 & 46.6 & 47.2 & 49.3 \\
& Fine-tuned (FP16) & 65.3 & 66.1 & 61.0 & 50.4 & 47.9 & 50.2 & 47.4 & 55.7 \\
\cmidrule(lr){2-10} 
& RTN (W4A4) & 50.5 & 50.6 & 41.8 & 26.5 & 25.5 & 35.9 & 24.3 & 37.0 \\
\gap
& GPTQ (W4A4) & 48.4 & 47.8 & 39.3 & 25.2 & 25.3 & 35.9 & 22.9 & 35.6 \\
\cmidrule(lr){2-10} 
& DuQuant (W4A4) & 54.6 & 68.1 & 51.4 & 46.6 & 45.2 & 44.6 & 48.0 & 51.5 \\
\gap
& QuaRot (W4A4) & 61.8 & 68.3 & 54.7 & 45.5 & 45.3 & 48.2 & 44.0 & 53.0 \\
\gap
& SpinQuant (W4A4) & 57.6 & 68.0 & 52.6 & 43.7 & 43.2 & 45.7 & 43.9 & 51.1 \\
\gap
& OSTQuant (W4A4) & 65.3 & 70.2 & 59.3 & 47.4 & 46.0 & 49.4 & 45.3 & 55.1 \\
\cmidrule(lr){2-10}
& \textbf{Ours (CAQ)} & 64.7 & 69.0 & 59.7 & 49.4 & 46.5 & 50.2 & 45.7 & \textbf{55.4} \\
\midrule
% Qwen2-7B Block
\multirow{8}{*}{\textbf{Qwen2-7B}} & Pre-trained (FP16) & 69.4 & 64.6 & 78.2 & 55.7 & 46.4 & 68.7 & 51.6 & 61.9 \\
 & Fine-tuned (FP16) & 72.7 & 66.6 & 77.1 & 71.5 & 63.3 & 71.6 & 64.2 & 69.4 \\
\cmidrule(lr){2-10} 
& RTN (W4A4) & 47.5 & 52.4 & 40.3 & 25.0 & 26.1 & 34.7 & 25.0 & 36.5 \\
\gap
& GPTQ (W4A4) & 48.5 & 52.9 & 39.2 & 25.9 & 26.7 & 36.4 & 23.7 & 36.9 \\
\cmidrule(lr){2-10} 
& QuaRot (W4A4) & 68.4 & 67.9 & 75.4 & 63.5 & 59.2 & 66.0 & 57.4 & 65.2 \\
\gap
& SpinQuant (W4A4) & 69.9 & 69.6 & 75.5 & 64.1 & 54.8 & 66.9 & 59.6 & 65.6 \\
\gap
& OSTQuant (W4A4) & 68.3 & 63.2 & 76.3 & 69.5 & 60.5 & 67.7 & 62.6 & 66.5 \\
\cmidrule(lr){2-10}
& \textbf{Ours (CAQ)} & 68.8 & 64.2 & 75.2 & 69.3 & 61.1 & 67.5 & 64.5 & \textbf{66.8} \\
\midrule
% Mistral-7B-v0.1 Block
\multirow{8}{*}{\textbf{Mistral-7B-v0.1}} & Pre-trained (FP16) & 57.8 & 55.1 & 62.9 & 60.4 & 53.9 & 49.4 & 57.4 & 56.1 \\
 & Fine-tuned (FP16) & 61.3 & 65.3 & 68.8 & 57.3 & 54.8 & 54.7 & 59.0 & 59.8 \\
\cmidrule(lr){2-10} 
& RTN (W4A4) & 52.7 & 49.8 & 40.8 & 29.4 & 25.4 & 37.2 & 28.9 & 38.3 \\
\gap
& GPTQ (W4A4) & 50.9 & 51.4 & 40.8 & 30.1 & 27.2 & 37.1 & 28.1 & 38.5 \\
\cmidrule(lr){2-10} 
& DuQuant (W4A4) & 60.1 & 67.5 & 56.3 & 49.9 & 51.8 & 53.9 & 52.6 & 56.4 \\
\gap
& QuaRot (W4A4) & 51.4 & 65.9 & 65.5 & 48.3 & 47.7 & 52.4 & 54.0 & 54.7 \\
\gap
& SpinQuant (W4A4) & 54.8 & 66.2 & 63.6 & 47.1 & 47.8 & 51.3 & 52.8 & 54.6 \\
\gap
& OSTQuant (W4A4) & 58.2 & 61.5 & 64.0 & 55.4 & 54.0 & 55.9 & 57.2 & 57.1 \\
\cmidrule(lr){2-10} 
& \textbf{Ours (CAQ)} & 61.5 & 64.0 & 66.0 & 54.3 & 52.0 & 53.7 & 54.7 & \textbf{57.8} \\
\bottomrule
\end{tabular}
\caption{
Category-wise SafetyBench accuracy (\%) for all W4A4 quantization methods across five models. This table details the results summarized in Table~\ref{tab:main_results}. The seven safety categories are: Offensiveness (OF), Unfairness and Bias (UB), Physical Health (PH), Mental Health (MH), Illegal Activities (IA), Ethics and Morality (EM), and Privacy and Property (PP). For each model, the best average safety score among quantized methods is \textbf{bolded}.
}
\label{tab:cat_safety}
\end{table*}

\section{Zero-shot Task Evaluation Results}
\label{sec:zero_shot_eval}

Table~\ref{tab:zero_shot_full_percentage} presents the zero-shot evaluation results on various benchmarks for LLaMA3.1-8B, LLaMA2-7B, LLaMA2-13B, Qwen2-7B, and Mistral-7B.
% We utilized 11 standard zero-shot tasks from \texttt{lm-evaluation-harness} to assess the general capabilities of the models:
All evaluations are conducted in a strict zero-shot setting following the \texttt{lm-evaluation-harness}~\cite{gao2021lm_eval_harness} protocol, with zero in-context examples and without any task-specific prompt tuning, calibration, or parameter updates.
We evaluate the models on 11 widely used benchmark tasks:

\begin{itemize}
    \item \textbf{ARC-c (AI2 Reasoning Challenge - Challenge Set)}~\cite{clark2018think}: Grade-school science questions requiring complex reasoning.
    \item \textbf{ARC-e (AI2 Reasoning Challenge - Easy Set)}~\cite{clark2018think}: Grade-school science questions.
    \item \textbf{BoolQ}~\cite{clark2019boolq}: Question answering with yes/no answers based on a passage.
    \item \textbf{CSQA (CommonsenseQA)}~\cite{talmor2019commonsenseqa}: Questions requiring commonsense knowledge.
    \item \textbf{HellaS. (HellaSwag)}~\cite{zellers2019hellaswag}: Completing sentences based on commonsense reasoning.
    \item \textbf{Lam. (LAMBADA)}~\cite{paperno2016lambada}: Predicting the last word of sentences in a passage.
    \item \textbf{MNLI (Multi-Genre Natural Language Inference)}~\cite{williams2018mnli}: Determining entailment, contradiction, or neutrality between sentence pairs.
    \item \textbf{OBQA (OpenBookQA)}~\cite{mihaylov2018obqa}: Answering questions using open book knowledge.
    \item \textbf{PIQA (Physical Interaction QA)}~\cite{bisk2020piqa}: Reasoning about physical interactions.
    \item \textbf{SIQA (Social Interaction QA)}~\cite{sap2019socialiqa}: Reasoning about social interactions.
    \item \textbf{WinoG. (Winogrande)}~\cite{sakaguchi2020winogrande}: Pronoun resolution problems requiring commonsense reasoning.
\end{itemize}

We compare CAQ with several baselines, including Pre-trained (FP16), Fine-tuned (FP16), RTN (W4A4), GPTQ (W4A4), DuQuant (W4A4), QuaRot (W4A4), SpinQuant (W4A4), and OSTQuant (W4A4). The results show that CAQ consistently achieves competitive or superior performance compared to other quantization methods across diverse models and tasks, demonstrating its effectiveness in maintaining general capabilities while ensuring safety alignment.

\section{Use of AI Assistants}
\label{sec:ai_assistants}
% We utilized AI assistants, specifically Gemini and ChatGPT, to aid in coding and writing during the preparation of this manuscript. The authors reviewed all AI-generated content and take full responsibility for the integrity of the work.

We acknowledge the use of AI assistants to aid in coding and writing during the preparation of this manuscript. Specifically, AI tools were utilized to refine the text and assist with code implementation. The authors reviewed and verified all AI-generated content and take full responsibility for the integrity of the work.

\begin{table*}[h!]
\centering
\resizebox{\textwidth}{!}{%
\begin{tabular}{llccccccccccccc}
\toprule
\textbf{Model} & \textbf{Method} & \textbf{PPL ($\downarrow$)} & \textbf{ARC-c} & \textbf{ARC-e} & \textbf{BoolQ} & \textbf{CSQA} & \textbf{HellaS.} & \textbf{Lam.} & \textbf{MNLI} & \textbf{OBQA} & \textbf{PIQA} & \textbf{SIQA} & \textbf{WinoG.} & \textbf{Avg. ($\uparrow$)} \\ \midrule
\multirow{9}{*}{\shortstack[l]{\textbf{LLaMA3.1-8B}}}
 & Pre-trained (FP16) & 6.14 & 51.71 & 81.40 & 82.11 & 71.50 & 59.98 & 75.37 & 49.79 & 33.20 & 80.14 & 47.03 & 74.03 & 67.24 \\
 & Fine-tuned (FP16) & 7.23 & 52.13 & 81.94 & 84.13 & 77.15 & 59.13 & 72.91 & 53.43 & 33.80 & 80.09 & 49.33 & 73.72 & 68.03 \\
 \cmidrule{2-15} 
 & RTN (W4A4) & 118.39 & 22.70 & 37.54 & 47.43 & 19.74 & 32.65 & 11.39 & 33.43 & 15.80 & 57.83 & 35.47 & 49.17 & 34.92 \\
 & GPTQ (W4A4) & 60.80 & 22.01 & 43.69 & 44.40 & 20.64 & 35.40 & 15.41 & 34.65 & 15.40 & 61.32 & 35.98 & 52.01 & 36.45 \\
 \cmidrule{2-15} 
 & DuQuant (W4A4) & 9.22 & 44.80 & 75.59 & 78.38 & 65.11 & 54.75 & 65.46 & 48.22 & 26.40 & 76.99 & 47.85 & 67.88 & 62.31 \\
 & QuaRot (W4A4) & 8.81 & 46.93 & 77.61 & 80.73 & 68.30 & 55.66 & 66.66 & 48.12 & 27.80 & 75.68 & 46.72 & 68.27 & 63.02 \\
 & SpinQuant (W4A4) & 8.57 & 47.35 & 79.00 & 82.23 & 71.83 & 55.74 & 68.99 & 50.15 & 29.00 & 77.04 & 48.11 & 68.82 & 64.57 \\
 & OSTQuant (W4A4) & 8.29 & 48.63 & 79.17 & 81.80 & 71.83 & 56.63 & 68.99 & 51.16 & 30.20 & 77.42 & 48.11 & 70.09 & 65.32 \\ 
 \cmidrule{2-15}
 & \textbf{Ours (CAQ)} & 8.41 & 48.81 & 80.09 & 82.97 & 71.42 & 56.91 & 70.37 & 50.45 & 32.00 & 77.64 & 47.75 & 71.03 & 65.54 \\ 
 \midrule
\multirow{9}{*}{\shortstack[l]{\textbf{LLaMA2-7B}}} 
 & Pre-trained (FP16) & 5.47 & 43.26 & 76.43 & 77.86 & 32.76 & 57.19 & 73.53 & 42.63 & 31.40 & 78.02 & 45.96 & 68.98 & 60.14 \\
 & Fine-tuned (FP16) & 6.94 & 44.20 & 73.91 & 79.85 & 58.56 & 57.78 & 70.75 & 49.68 & 32.80 & 76.33 & 49.08 & 66.38 & 62.32 \\ 
 \cmidrule{2-15} 
 & RTN (W4A4) & 868.46 & 21.08 & 27.90 & 53.91 & 20.80 & 27.22 & 1.92 & 35.13 & 16.80 & 53.86 & 34.54 & 49.80 & 32.56 \\
 & GPTQ (W4A4) & 3251.19 & 20.65 & 28.24 & 56.73 & 19.00 & 27.37 & 4.50 & 34.90 & 13.20 & 53.21 & 34.85 & 48.86 & 32.56 \\
 \cmidrule{2-15} 
 & DuQuant (W4A4) & 7.88 & 41.13 & 70.16 & 76.88 & 52.58 & 54.32 & 67.24 & 49.49 & 29.80 & 74.70 & 45.60 & 63.93 & 59.20 \\
 & QuaRot (W4A4) & 7.67 & 39.08 & 69.82 & 76.82 & 49.14 & 54.59 & 67.84 & 44.42 & 29.40 & 75.08 & 46.83 & 64.80 & 58.36 \\
 & SpinQuant (W4A4) & 7.68 & 40.36 & 70.16 & 77.16 & 50.70 & 54.84 & 67.48 & 47.57 & 29.40 & 73.99 & 45.96 & 64.33 & 58.61 \\
 & OSTQuant (W4A4) & 7.28 & 42.06 & 71.72 & 78.29 & 52.42 & 55.35 & 69.69 & 49.64 & 31.20 & 74.21 & 47.03 & 65.19 & 60.17 \\ 
 \cmidrule{2-15}
 & \textbf{Ours (CAQ)} & 7.56 & 40.10 & 69.82 & 76.97 & 48.89 & 55.12 & 68.83 & 43.20 & 31.40 & 75.52 & 46.42 & 65.90 & 59.05 \\
 \midrule
\multirow{9}{*}{\shortstack[l]{\textbf{LLaMA2-13B}}}
 & Pre-trained (FP16) & 4.88 & 48.21 & 79.55 & 80.49 & 47.01 & 60.11 & 76.71 & 43.29 & 35.00 & 79.16 & 47.19 & 71.82 & 63.53 \\
 & Fine-tuned (FP16) & 6.11 & 46.25 & 77.57 & 81.68 & 64.54 & 60.72 & 73.04 & 47.54 & 35.60 & 77.80 & 50.15 & 71.11 & 65.04 \\
 \cmidrule{2-15} 
 & RTN (W4A4) & 4517.82 & 19.88 & 28.03 & 39.24 & 19.16 & 26.26 & 0.06 & 34.59 & 12.80 & 55.77 & 35.16 & 50.67 & 30.72 \\
 & GPTQ (W4A4) & 3642.56 & 21.33 & 29.55 & 39.08 & 18.84 & 25.93 & 0.04 & 35.49 & 13.40 & 53.54 & 34.08 & 47.51 & 30.03 \\
 \cmidrule{2-15} 
 & DuQuant (W4A4) & 6.66 & 44.28 & 76.22 & 77.68 & 59.38 & 58.75 & 70.41 & 47.81 & 34.00 & 75.19 & 49.08 & 66.22 & 62.33 \\
 & QuaRot (W4A4) & 6.73 & 42.24 & 75.00 & 76.85 & 58.56 & 57.71 & 70.66 & 44.65 & 31.60 & 75.95 & 48.72 & 69.14 & 61.91 \\
 & SpinQuant (W4A4) & 6.62 & 43.52 & 74.96 & 75.26 & 6143 & 58.11 & 70.89 & 45.35 & 32.60 & 76.71 & 49.64 & 68.19 & 62.55 \\
 & OSTQuant (W4A4) & 6.32 & 43.34 & 76.05 & 79.02 & 60.11 & 58.91 & 71.94 & 49.40 & 32.40 & 76.12 & 49.59 & 69.53 & 63.16 \\
 \cmidrule{2-15}
 & \textbf{Ours (CAQ)} & 6.90 & 43.52 & 75.80 & 81.10 & 62.16 & 59.15 & 71.59 & 48.39 & 34.60 & 76.06 & 49.95 & 71.27 & 63.56 \\
 \midrule
\multirow{9}{*}{\shortstack[l]{\textbf{Qwen2-7B}}} 
 & Pre-trained (FP16) & 7.13 & 48.46 & 79.25 & 84.74 & 81.57 & 59.37 & 71.80 & 57.75 & 35.20 & 79.87 & 48.46 & 72.30 & 67.82 \\
 & Fine-tuned (FP16) & 7.60 & 51.11 & 80.43 & 85.20 & 81.16 & 60.99 & 70.00 & 58.77 & 34.80 & 79.60 & 52.25 & 69.85 & 68.66 \\
 \cmidrule{2-15} 
 & RTN (W4A4) & 5453.45 & 20.82 & 26.89 & 37.89 & 20.15 & 25.92 & 0.04 & 33.74 & 14.40 & 52.67 & 34.14 & 48.38 & 30.28 \\
 & GPTQ (W4A4) & 5024.77 & 20.90 & 27.06 & 37.98 & 19.08 & 26.01 & 0.00 & 34.50 & 14.80 & 53.59 & 34.19 & 48.70 & 29.86 \\
 \cmidrule{2-15} 
 & QuaRot (W4A4) & 8.43 & 48.81 & 77.53 & 82.60 & 77.07 & 57.94 & 65.01 & 54.70 & 33.20 & 76.77 & 49.74 & 66.22 & 65.69 \\
 & SpinQuant (W4A4) & 8.41 & 49.32 & 78.24 & 83.98 & 77.40 & 57.97 & 66.31 & 55.89 & 31.60 & 77.64 & 49.33 & 66.85 & 65.94 \\
 & OSTQuant (W4A4) & 8.18 & 51.19 & 78.58 & 85.54 & 77.81 & 58.90 & 66.35 & 57.42 & 35.60 & 76.77 & 51.59 & 67.09 & 67.01 \\
 \cmidrule{2-15}
 & \textbf{Ours (CAQ)} & 8.23 & 50.26 & 78.20 & 84.56 & 77.23 & 58.49 & 65.55 & 56.24 & 34.20 & 77.86 & 50.51 & 68.51 & 66.37 \\
 \midrule
\multirow{9}{*}{\shortstack[l]{\textbf{Mistral-7B-v0.1}}} 
 & Pre-trained (FP16) & 5.25 & 50.26 & 80.72 & 83.70 & 56.43 & 61.23 & 75.65 & 45.49 & 32.80 & 80.52 & 46.67 & 73.72 & 65.65 \\
 & Fine-tuned (FP16) & 6.84 & 50.09 & 79.97 & 80.06 & 64.13 & 56.28 & 69.20 & 48.73 & 32.00 & 79.38 & 50.46 & 69.69 & 64.35 \\
 \cmidrule{2-15} 
 & RTN (W4A4) & 40.59 & 30.03 & 56.52 & 52.94 & 23.26 & 39.22 & 23.91 & 35.40 & 20.20 & 645.8 & 39.25 & 52.41 & 41.46 \\
 & GPTQ (W4A4) & 30.55 & 31.14 & 59.26 & 53.21 & 25.80 & 37.00 & 28.99 & 35.57 & 22.40 & 66.38 & 40.38 & 53.83 & 43.11 \\
 \cmidrule{2-15} 
 & DuQuant (W4A4) & 7.20 & 48.38 & 78.70 & 78.44 & 61.75 & 54.98 & 67.05 & 42.67 & 31.20 & 77.75 & 49.33 & 68.90 & 62.49 \\
 & QuaRot (W4A4) & 7.09 & 47.95 & 79.25 & 78.65 & 62.33 & 55.12 & 67.92 & 46.81 & 29.40 & 77.42 & 49.39 & 67.72 & 62.83 \\
 & SpinQuant (W4A4) & 7.08 & 49.06 & 79.00 & 79.20 & 62.90 & 54.96 & 67.44 & 47.05 & 30.20 & 78.02 & 49.08 & 68.11 & 63.10 \\
 & OSTQuant (W4A4) & 7.07 & 48.38 & 79.25 & 79.85 & 62.00 & 55.23 & 67.81 & 47.81 & 31.00 & 77.20 & 50.26 & 69.85 & 63.28 \\
 \cmidrule{2-15}
 & \textbf{Ours (CAQ)} & 7.42 & 48.89 & 79.00 & 78.53 & 63.06 & 55.18 & 67.65 & 46.93 & 30.20 & 77.53 & 49.18 & 68.19 & 62.93 \\
 \bottomrule
\end{tabular}%
}
\caption{Zero-shot evaluation results (Accuracy, \%) and Perplexity (PPL). We report PPL, accuracy for each task, and the average score. CAQ (Ours) consistently demonstrates competitive or superior performance compared to other quantization methods across diverse models.}
\label{tab:zero_shot_full_percentage}
\end{table*}

\end{document}

%% file: 1_intro.tex
\section{Introduction}

Post-Training Quantization (PTQ) has emerged as the de-facto standard for efficient LLM deployment, enabling significant reductions in memory and latency by representing weights and activations in low precision (\eg 4-bit) without expensive retraining. A growing body of work has advanced PTQ through increasingly sophisticated pre-quantization transformations~\cite{xiao2023smoothquant, ashkboos2024quarot, lin2024duquant, sun2024flatquant}, all sharing a common optimization target: minimizing reconstruction error (\eg MSE or KL divergence) between the full-precision and quantized outputs. While effective at preserving perplexity and task accuracy, we ask a critical question that has been largely overlooked: \textit{is reconstruction error a sufficient PTQ objective for deployment-ready models?}

We argue that it is not. Modern LLMs are fine-tuned through alignment procedures such as reinforcement learning from human feedback (RLHF)~\cite{ouyang2022training, bai2022helpful}, which instill safety behaviors that are not captured by reconstruction metrics. Recent findings expose this objective mismatch as a concrete vulnerability: \citet{kharinaev2025impact} demonstrate that PTQ can silently erase safety guardrails instilled by RLHF—causing models to revert to unsafe behaviors despite maintaining low perplexity—a phenomenon they term \textit{alignment degradation}. In response, \citet{chen2025qresafe} attempt to recover alignment via post-hoc fine-tuning of already-quantized models. However, patching the output without addressing the objective itself is fundamentally limited, requiring additional curated data and forfeiting the efficiency gains PTQ is designed to provide.

To address this root cause, we propose \textbf{Contrastive Alignment Quantization (CAQ)}, which extends the PTQ objective design space beyond reconstruction fidelity by integrating a \textbf{Contrastive Alignment Loss (CAL)} directly into the optimization process. CAL implements a principled push-pull mechanism: it pulls the quantized model toward the safe, fine-tuned reference while pushing it away from the unaligned, pre-trained model. Our approach follows the transformation-based PTQ paradigm, allowing CAL to seamlessly integrate into existing pipelines with only a small, unlabeled calibration set—requiring no specialized safety datasets and introducing negligible computational overhead (See Figure~\ref{fig:example}).

Our extensive experiments demonstrate that CAQ enables robust 4-bit (W4A4) quantization across diverse LLM architectures—including LLaMA2~\citep{touvron2023llama2}, LLaMA3.1~\citep{grattafiori2024llama3}, Qwen2~\citep{yang2024qwen2technicalreport}, and Mistral~\citep{jiang2023mistral7b}—achieving superior safety alignment where state-of-the-art PTQ methods fail, without sacrificing general capabilities. To the best of our knowledge, CAQ is the first framework to demonstrate that a well-designed PTQ objective can simultaneously achieve quantization fidelity and behavioral alignment within a standard PTQ pipeline.

Our contributions are summarized as follows:
\begin{itemize}
    \item We identify reconstruction error as an incomplete PTQ objective and propose \textbf{CAQ} as the first principled framework that extends PTQ optimization to jointly preserve quantization fidelity and behavioral alignment.
    % \vspace{-1mm}
    \item We introduce the \textbf{Contrastive Alignment Loss (CAL)}, a generalizable PTQ objective that incorporates alignment signals via a contrastive push-pull mechanism, requiring no specialized datasets and integrating seamlessly into existing transformation-based PTQ pipelines.
    % \vspace{-1mm}
    \item Through experiments on diverse model architectures, we demonstrate that CAQ consistently achieves best-in-class safety alignment at W4A4 quantization with negligible overhead, validating its role as a practical and generalizable framework for deployment-ready quantization.
\end{itemize}

%% file: 2_relwork.tex
\section{Related Work}

\subsection{Post-Training Quantization for LLMs}

Quantization maps high-precision values to a lower bit-width, reducing the memory and computation footprint of LLMs. It is now the de-facto standard for efficient LLM deployment. Post-training quantization (PTQ) methods such as GPTQ~\cite{frantar2023gptq} and AWQ~\cite{lin2024awq} achieve efficiency without full retraining. Recent work focuses on the Weight-4-bit, Activation-4-bit (W4A4) setting, which is highly efficient yet susceptible to performance degradation.

The quantization process can be formulated as:
\vspace{-1mm}
\begin{equation}
\label{eq:minmax}
    Q(\mathbf{X}) = s \cdot \text{round}\left(\frac{\mathbf{X} - z}{s}\right) + z ,
\end{equation}
where $s$ and $z$ are the quantization parameters, known as the scale and zero-point, respectively. Here, $\mathbf{X}$ and $Q(\mathbf{X})$ are the full-precision and quantized tensors, and $N$ is the number of bits.

The performance of a quantized model is typically evaluated by calculating quantization error. This error is defined as the difference between the output of the full-precision layer and the quantized layer, often referred to as the reconstruction error. For a full-precision output $\mathbf{Y}=\mathbf{W}\mathbf{X}$ and its quantized counterpart $\mathbf{Y}_Q=Q(\mathbf{W})Q(\mathbf{X})$, the error is formulated as the mean squared error:
\vspace{-1mm}
\begin{equation}
\label{eq:quant_error}
    E_{\text{quant}} = \| \mathbf{Y} - \mathbf{Y}_Q \|_F^2 = \| \mathbf{W}\mathbf{X} - Q(\mathbf{W})Q(\mathbf{X}) \|_F^2,
\end{equation}
where $\|\cdot\|_F$ denotes the Frobenius norm.

To mitigate quantization errors, recent works apply \textit{equivalent transformation} to weights and activations before quantization. Such methods reshape the distributions by multiplying a transformation matrix $\mathbf{T}$ and its inverse $\mathbf{T}^{-1}$ as follows:
\vspace{-1mm}
\begin{equation}
\mathbf{Y} = \mathbf{W}\mathbf{X} = (\mathbf{W}\mathbf{T})(\mathbf{T}^{-1}\mathbf{X})
\end{equation}
The matrix $\mathbf{T}$ is determined to make the distributions more amenable to quantization, analytically or through optimization. Crucially, this structure preserves the exact output of the full-precision model. Furthermore, no additional computational overhead occurs for the linear layer operations during inference since the transformation matrix can be fused with the original weight.
Prominent examples of this paradigm include SmoothQuant~\cite{xiao2023smoothquant}, which scales down outlier activations, and QuaRot~\cite{ashkboos2024quarot} and SpinQuant~\cite{liu2025spinquant}, which apply rotation-based transformations. DuQuant~\cite{lin2024duquant}, FlatQuant~\cite{sun2024flatquant}, and OstQuant~\cite{hu2025ostquant} introduce adaptive transformations that can effectively handle outliers and improve model performance after quantization. 
% While these methods are effective at preserving perplexity and accuracy by optimizing reconstruction losses (\eg MSE, KL), their objectives are fundamentally unaware of the fine-grained alignment behaviors introduced through fine-tuning, such as RLHF.
While these methods are effective at preserving perplexity and accuracy, their reconstruction-based objectives define an incomplete optimization target: they minimize distributional deviation at the output level, but remain blind to behavioral alignment—a property instilled through fine-tuning that cannot be recovered from reconstruction error alone.

\subsection{Behavioral Alignment as an Overlooked PTQ Objective}

Significant effort has been devoted to aligning LLMs with human values through instruction tuning and reinforcement learning from human feedback (RLHF)~\cite{ouyang2022training, bai2022helpful}. While the community has pursued LLM alignment and quantization as largely independent goals, recent findings reveal a dangerous intersection. \citet{kharinaev2025impact} first discovered that PTQ can silently erase safety guardrails instilled by RLHF—causing models to revert to unsafe behaviors despite maintaining low perplexity—a phenomenon they term \textit{alignment degradation}. \citet{dong2025qmisalign} further show that certain safety vulnerabilities remain dormant in full-precision models but become exposed only after 4-bit quantization, while \citet{zhang2025unlearningfail} demonstrate that models can recover up to 83\% of previously unlearned sensitive information post-quantization. Together, these findings expose alignment degradation as a direct consequence of the incomplete PTQ objective: reconstruction error provides no signal to prevent behavioral regression.

In response, Q-resafe~\cite{chen2025qresafe} introduces a post-hoc patching framework that attempts to recover alignment in an already-quantized model. However, patching the output without addressing the objective itself is fundamentally limited—it requires additional curated data and is restricted to weight-only, mixed-precision quantization, forfeiting the efficiency gains PTQ is designed to provide. This motivates a forward-compatible solution that integrates alignment preservation directly into the quantization objective.

\subsection{Positioning Our Work}

Our work extends the PTQ objective design space beyond reconstruction fidelity. While prior methods optimize MSE or KL divergence to minimize quantization error, CAL introduces a contrastive objective that jointly captures distributional fidelity and behavioral alignment—offering a more complete optimization target for deployment-ready quantized models. Unlike post-hoc approaches, CAQ integrates alignment preservation directly into the transformation optimization step, requiring no separate corrective fine-tuning stages. CAQ is optimized with a general-purpose calibration set common in PTQ research, requires no specialized safety datasets, and introduces negligible computational overhead—making it fully compatible with standard PTQ pipelines.